\documentclass[review]{elsarticle}

\usepackage{geometry}

\usepackage{lineno,hyperref}
\modulolinenumbers[1]
\hypersetup{colorlinks=true, citecolor=blue}
\usepackage{gensymb} 
\usepackage{amsmath} 
\usepackage{graphics} 
\graphicspath{{./figures/}}
\usepackage{booktabs} 
\usepackage{makecell}
\usepackage{array}
\usepackage{caption}
\renewcommand{\arraystretch}{2}

\usepackage{setspace}

\usepackage{textcomp}
\usepackage{upgreek}
\usepackage{amsmath}
\usepackage{amssymb}

\usepackage{array, booktabs, multirow, longtable}
\usepackage{adjustbox}
\usepackage{pdflscape}
\usepackage{threeparttablex}
\usepackage{microtype}

\usepackage{a4wide}

\usepackage{bm}

\journal{}

\biboptions{authoryear}

\usepackage{cancel}
\begin{document}
\newgeometry{top=2cm, bottom=2cm}

\begin{frontmatter}

\title{From physical surfaces to human-centric heat stress: LST and UTCI heat mapping reveals nonlinear effects of urban morphology}

\author[addr1,addr2]{Yuan Wang}
\author[addr4]{Shengao Yi}
\author[addr4]{Xiaojiang Li}
\author[addr5]{Pengyuan Liu}
\author[addr6]{Zhiwei Yang}
\author[addr2,addr3]{Ronita Bardhan}
\author[addr1,addr2]{Rudi Stouffs\corref{correspondingauthor}}
\cortext[correspondingauthor]{Correspondence to: Rudi Stouffs (stouffs@nus.edu.sg)}

\address[addr1]{Department of Architecture, National University of Singapore, Singapore 117566, Singapore}
\address[addr2]{Cambridge Centre for Advanced Research and Education in Singapore (CARES), Singapore 138602, Singapore}
\address[addr3]{Sustainable Design Group, Department of Architecture, University of Cambridge, Cambridge, United Kingdom}
\address[addr4]{Department of City and Regional Planning, University of Pennsylvania, Philadelphia, PA 19104, USA}
\address[addr5]{Urban Analytics Subject Group, Urban Studies \& Social Policy Division, University of Glasgow}
\address[addr6]{Laboratory for Earth Surface Processes, Ministry of Education, College of Urban and Environmental Sciences, Peking University, Beijing 100871, China}

\begin{abstract}

Heat exposure connects the built environment and public health, directly shaping the livability and sustainability of urban areas. Fully understanding the spatial heterogeneity of heat exposure and its driving factors is therefore a vital prerequisite for climate-adaptive urban planning. However, most planning-oriented studies rely on land surface temperature (LST), and whether LST adequately represents human heat exposure and how it differs from physiologically relevant heat stress remains insufficiently examined. Here, adopting Landsat-retrieved 30-m LST and GPU-accelerated 1-m universal thermal climate index (UTCI) in Singapore, this study establishes a comprehensive ``Modeling-Comparing-Assessing" framework to systematically evaluate the spatial and mechanistic discrepancies between the two metrics. We further investigate pronounced non-stationary and threshold-based quantitative relationships of the two metrics with urban factors by employing a novel geographically weighted XGBoost (GW-XGBoost) and generalized additive model (GAM) workflow. Our results demonstrate notable discrepancies in spatial patterns of LST and UTCI, along with substantial spatial heterogeneity in how 2D and 3D urban factors impact these two thermal metrics, as revealed by explainable GW-XGBoost models (the test R$^2$ = 0.855 for LST and 0.905 for UTCI, respectively). Crucially, spatially explicit SHAP interprets that sky view factor plays a central role in explaining UTCI variability but exhibits a comparatively marginal independent contribution to LST, indicating that LST inadequately captures shading-driven and radiative processes governing actual human heat stress. Notably, SHAP-GAM analysis indicates that higher albedo is associated with increased UTCI. These novel findings provide model-informed planning implications for integrating physiologically relevant thermal indices to support targeted heat risk management and climate-adaptive urban planning.

\end{abstract}


\end{frontmatter}

\clearpage

\doublespacing

\section{Introduction}
Climate-informed urban planning increasingly depends on understanding how urban morphology shapes the local thermal environment. Previous research has established strong associations between urban form and thermal conditions through parameters such as sky view factor \citep{oke2002boundary,eliasson1996urban,zakvsek2011sky,robinson2006urban}, building density \citep{yao2023exploring,shi2019assessing}, and land cover \citep{chen2024impacts,zhang2024spatiotemporal,li2024multidisciplinary,chang2023monitoring}. However, this body of work overwhelmingly uses land surface temperature (LST) derived from satellite observations as the dependent variable, owing to the fine spatial resolution and long-term temporal coverage of LST products. Such a reliance on LST constitutes a fundamental methodological limitation rather than a neutral proxy choice. Although LST offers long-term, spatially continuous coverage and computational convenience \citep{wang2024integrated}, it measures radiometric surface heating of roofs and treetops and therefore fails to capture the radiative exchange, convective cooling, humidity effects, and wind modulation that jointly determine human heat stress within street canyons \citep{zhan2025satellite}. These omissions are particularly consequential in high-density urban environments, where shading geometry, enclosure effects, and vegetation modify the radiative and convective environment in ways that cannot be inferred from surface temperatures alone \citep{briegel2025satellite,zhan2025satellite}. 

More fundamentally, LST captures only the heat component and excludes the humidity, wind, and multidirectional radiation that determine physiologically relevant heat stress. Existing heat-exposure models that use LST as a proxy may achieve high predictive performance while remaining poorly aligned with human physiological processes \citep{nazarian2022integrated,tuholske2021global}. As a result, LST-based assessments routinely mischaracterize human-centered heat stress, which reflects physiological heat strain, and can lead to biased energy demand estimates and inappropriate adaptation strategies. This persistent mismatch between surface-based temperature indicators and human-centered thermal conditions exposes a critical gap in current urban morphology–climate frameworks, which lack scalable approaches capable of representing human-centered thermal stress at the city scale \citep{nazarian2022integrated}. Addressing this gap requires moving beyond surface-based temperature indicators towards models that explicitly resolve the spatial complexity of the heat stress and systematically identify where—and why—LST fails as a decision-relevant metric \citep{zhan2025satellite,muse2024daytime,nazarian2022integrated}.

Human heat stress proxies such as the Universal Thermal Climate Index (UTCI) address this gap by integrating air temperature ($T_a$), humidity, the mean radiant temperature ($T_{mrt}$), and wind speed \citep{brode2012deriving,fiala2012utci,jendritzky2012utci,blazejczyk2012comparison}. UTCI is particularly valuable because of its thermophysiological basis, international standardization, and stable sensitivity across climates. Its accuracy depends on the estimation of $T_{mrt}$, which captures the full shortwave and longwave radiative environment surrounding the human body. To compute this at scale, physically based models such as ENVI met \citep{roth2017evaluation}, RayMan \citep{matzarakis2007modelling}, and SOLWEIG (SOlar and LongWave Environmental Irradiance Geometry \citep{lindberg2008solweig,lindberg2011influence,lindberg2018urban}) have been developed. Among these, SOLWEIG offers a practical balance of physical realism and computational efficiency by explicitly representing six directional radiation components. Recent advances in GPU acceleration now make it possible to apply SOLWEIG citywide at fine spatial resolution \citep{li2021gpu,yi2025hyperlocal}.

Although simulation models provide physically realistic estimates, understanding where and how urban morphology drives spatial variations in thermal conditions requires complementary statistical modeling. Crucially, the necessity for advanced modeling arises from the fundamental physical properties of urban microclimates. Theoretically, the relationships between urban form and thermal environment are inherently non-linear and spatially non-stationary \citep{wan2025exploring,zhou2025lcz,zhao2026incorporating}. These non-linear relationships are driven by competing microclimatic mechanisms in built environments. For example, compact urban forms can increase pedestrian-level thermal stress by intensifying heat storage and obstructing ventilation, whereas a lower sky view factor (SVF) can reduce daytime shortwave exposure through shading. Consistent with this trade-off, \citet{he2015influence} showed that highly shaded areas (SVF $<$ 0.3) experienced fewer hot conditions in summer but longer periods of cold discomfort in winter than moderately shaded areas (0.3 $<$ SVF $<$ 0.5) and slightly shaded areas (SVF $>$ 0.5). This finding suggests that the relationship between SVF and thermal comfort is context-dependent rather than strictly monotonic. Similarly, canopy cover can substantially reduce thermal stress through shading and evapotranspiration \citep{li2024cooling,yu2024enhanced}. However, tree cooling effects can be nonlinearly amplified under rising background temperatures, but may also be reduced or even reversed under certain conditions due to stomatal closure, longwave radiation trapping, and increased aerodynamic resistance \citep{chen2026machine}. Their efficacy therefore varies across regions depending on tree traits, urban morphology, and climatic conditions \citep{li2024cooling}. However, existing studies mainly rely on linear regression \citep{zhang2023multi}, spatial regression \citep{wongsai2024spatial,assaf2024modeling}, or machine learning (ML) models \citep{zhang2024influence,yang2024optimizing,chen2023constructing} to examine these relationships. Linear models cannot represent nonlinear interactions and threshold effects, and both linear and conventional ML models typically ignore spatial heterogeneity. Spatial regression models such as geographically weighted regression (GWR) account for local variation but are limited by their linear structure and susceptibility to multicollinearity. Recent developments in geographically weighted machine learning address these limitations by combining nonlinear modeling with spatial weighting \citep{fouedjio2024locally,yang2023geographically,wang2024geographically,yi2024interpretable}. The geographically weighted XGBoost (GW-XGBoost) model is particularly up-and-coming because it captures nonlinear interactions, accommodates high dimensional datasets, and reveals local variable importance while maintaining interpretability through SHapley Additive exPlanations (SHAP) analysis. Characterizing the spatial heterogeneity of these geographical patterns is essential for mapping disparities in heat exposure impacts, identifying hotspot areas, and informing the design of targeted local and national interventions to mitigate current and potential future risks.

Beyond methodological considerations, a more fundamental question remains insufficiently addressed: whether commonly used surface-based temperature metrics are sufficient to inform human-centric heat mitigation and urban planning decisions, which prioritize human-related parameters over purely physical indicators \citep{yang2025human}. LST primarily reflects surface thermal properties rather than the atmospheric and radiative conditions experienced by humans. As a result, its relevance for assessing heat stress, particularly in dense and vertically complex urban environments, remains uncertain. In contrast, UTCI is explicitly designed to represent human thermal stress. However, despite its growing use, it remains unclear whether UTCI is systematically more sensitive to urban spatial structure and therefore more informative for place-based, human-centric decision-making. Addressing this gap is critical for evaluating the appropriateness and limitations of temperature metrics used in urban heat assessment.

This study aims to address the fundamental gaps between surface-based and human-centered thermal indicators by applying a GW-XGBoost framework to jointly analyze LST and UTCI. This approach accounts for spatial heterogeneity and correlation, enabling a detailed comparison of where and why LST diverges from UTCI in high-density tropical coastal cities under strong solar radiation, using Singapore as a case study. The study is guided by three research questions:

(1) What and where are the differences in LST and UTCI?

(2) How do complex 2D and 3D urban morphological features drive these discrepancies through non-linear and spatially non-stationary mechanisms?

(3) Which specific urban factors are systematically misrepresented when relying solely on LST for climate-adaptive planning?

\section{Methodology}

This study employs a high-resolution comparative framework in Singapore, integrating Landsat-retrieved LST and GPU-accelerated 1-m UTCI with multi-source 2D and 3D urban covariates. By implementing GW-XGBoost and SHAP-based interpretation, we quantify the spatial non-stationarity and divergent driving mechanisms of these two thermal metrics. 

\subsection{Study area}
\begin{figure}[h!]
    \centering   \includegraphics[width=\linewidth]{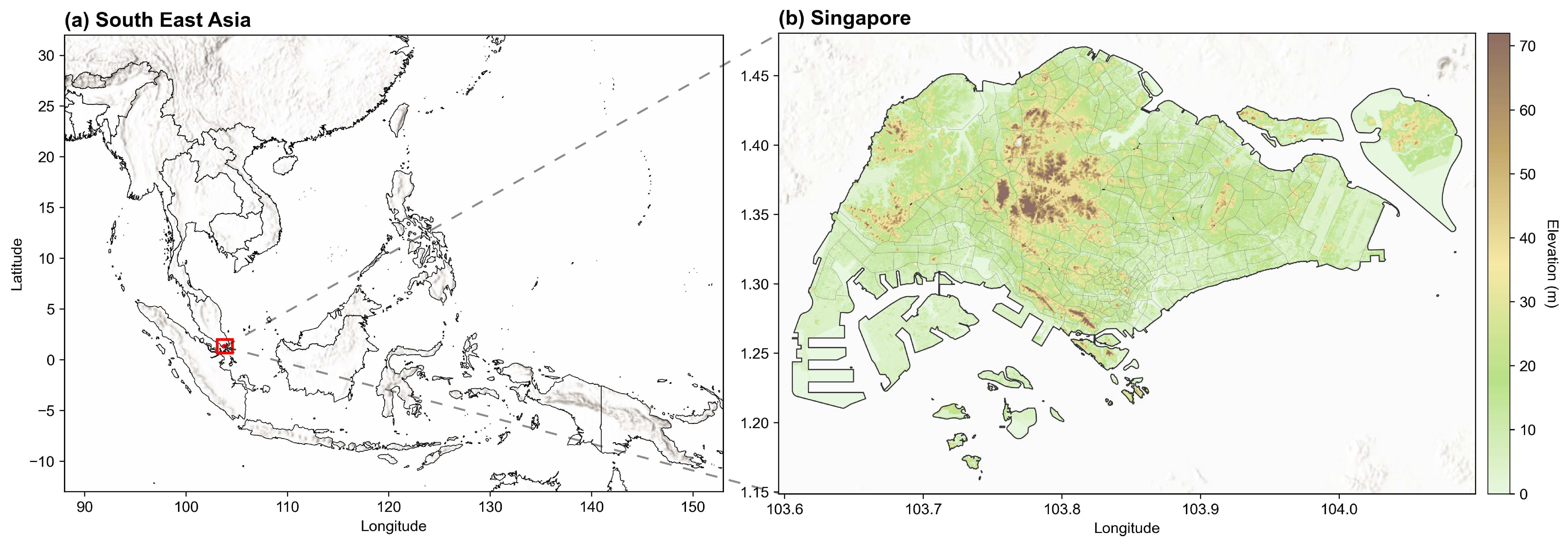}
    \caption{The regional location of Singapore. (a) Location of Singapore (red box) relative to the broader Southeast Asian region. (b) Topographical map of Singapore displaying a Digital Elevation Model (DEM) alongside urban subzone boundaries.}
    \label{fig:SG_region}
\end{figure}

\begin{figure}[h!]
    \centering   \includegraphics[width=1.0\linewidth]{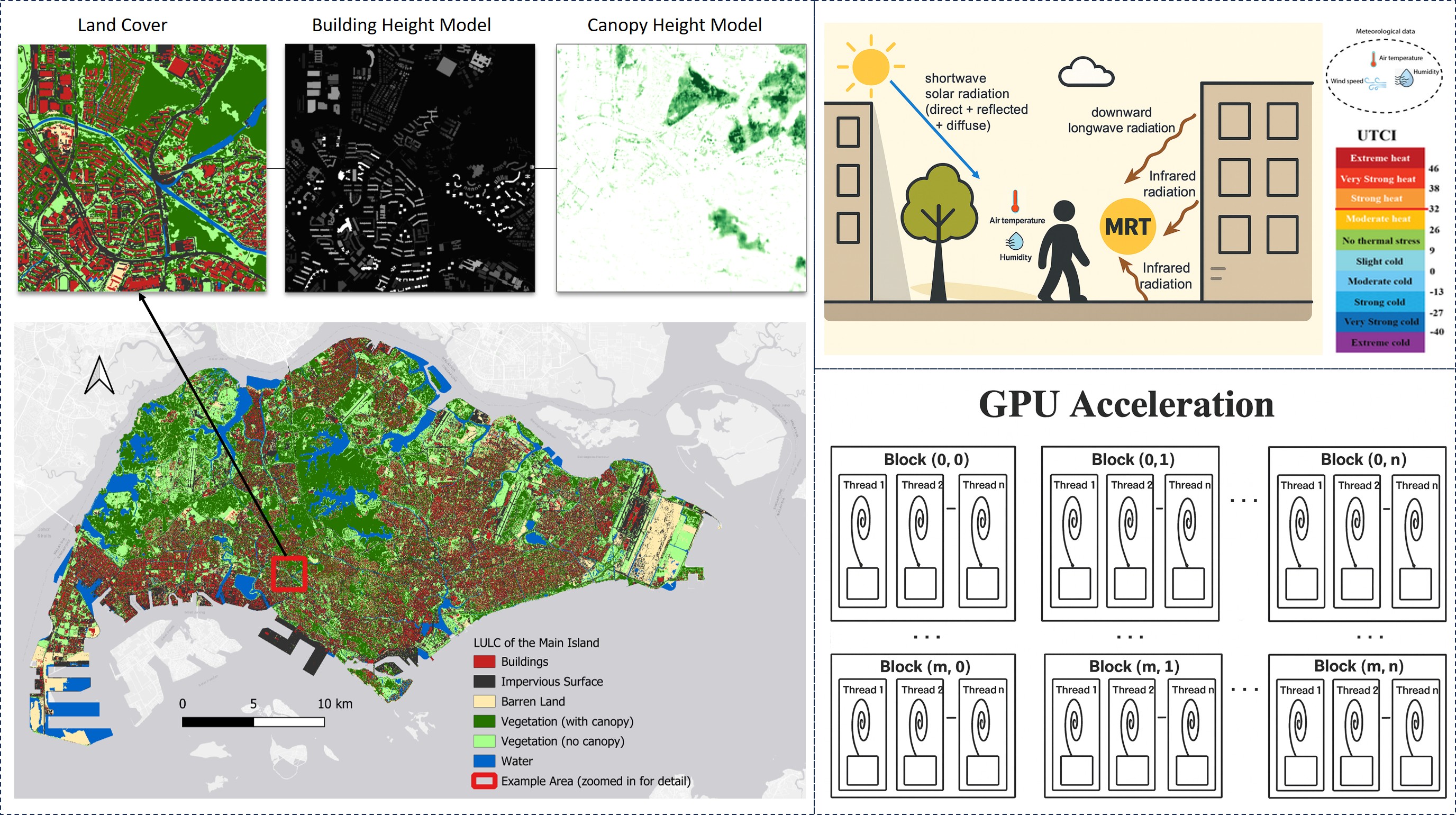}
    \caption{Singapore mainland with 1-m land use/land cover (LULC), building height, and canopy height layers, and the workflow for $T_{mrt}$ and UTCI computation using SOLWEIG and the human physiological model integrated in GPU-accelerated pipelines. The area marked in red is shown as an illustrative inset to highlight high-resolution spatial details that are imperceptible at the full island scale.}
    \label{fig:utci}
\end{figure}

Singapore, a high-density city-state in Southeast Asia (Figure~\ref{fig:SG_region}), provides an ideal natural laboratory for examining how urban morphology shapes the spatial heterogeneity of LST and UTCI. Despite its small land area, the city exhibits pronounced morphological contrasts arising from intense vertical development, diverse land-use configurations, and fine-grained variations in building height, density, and vegetation structure (Figure~\ref{fig:utci}). These characteristics are embedded within a tropical equatorial climate with persistently high temperatures and humidity and minimal seasonal variability, allowing observed thermal differences to be primarily attributed to urban form.

Singapore’s urban fabric comprises a heterogeneous mix of high-rise residential estates, dense commercial cores, industrial zones, and an extensive, strategically distributed network of green infrastructure, including parks, urban forests, roadside vegetation, vertical greenery, and rooftop gardens. This combination of strong 3D morphological variability and climate stability makes Singapore particularly well suited for isolating and quantifying the differential influences of 2D and 3D urban form on LST and UTCI. Moreover, the availability of high-resolution urban morphology and meteorological datasets enables the application of interpretable geographically weighted machine learning models to capture spatially varying morphology–temperature relationships.

\subsection{Variables and Data Processing}

Two target temperature variables LST and UTCI, are shown in Table \ref{tab:varibles}. LST characterizes the radiometric skin temperature of the land surface, and UTCI integrates $T_a$, $T_{mrt}$, humidity, and wind speed to reflect human heat stress. May 2019 was selected as the study period because May is typically among the hottest months of the year in Singapore, and 2019 ranks as one of the warmest years on record \citep{MSS2023climate,MSS2020annual}. The hottest climatic conditions enhance the contrast in LST and UTCI, providing an ideal context for examining urban heat patterns and spatial variability in human heat exposure across the city.

\begin{landscape}
\begin{table}[htpb]
\captionsetup{singlelinecheck=false}
\vspace*{-1cm}
\caption{Summary of variables, formulae, units, descriptions, and references of multi-source datasets.}
\resizebox{\linewidth}{!}{
\begin{threeparttable}
\renewcommand{\arraystretch}{1.20}
\begin{tabular}{lllllll}
\hline
\textbf{Category} & \textbf{Variables} & \textbf{Formula} & \textbf{Unit} & \textbf{Description } &  \textbf{Reference}\\ \hline

LST & Landsat LST & Described in the main text & °C & daytime $\sim$11:00 a.m. SGT, 16-day repeat cycle, 30-m & \citet{wang2020dynamic}\\ 

UTCI & UTCI & Described in the main text & °C & daytime, 11:00 a.m. SGT, hourly, 1-m & \makecell[l]{\parbox[t]{3.5cm}{\citet{lindberg2008solweig,brode2012deriving}}}\\ 

3D urban morphology & SVF & \makecell[l]{
$\mathrm{SVF} = 
\sum_{i=1}^{n} 
S \,
\frac{2}{2\pi}
\sin\!\left(\frac{\pi}{180}\right)
\sin\!\left(\frac{\pi(2\alpha_i - 1)}{2n}\right)
\frac{360}{\theta_i}$)
} & -- & \makecell[l]{Annulus-weighted SVF, ratio of visible sky hemisphere,\\ indicating proportion of hemispheric radiation accessible \\to a planar surface.} & \citet{lindberg2010continuous} \\ 
  & BH & $ \frac{\sum_{i=1}^{n} \text{BH}_i}{n}$ & m & Average building height in the spatial unit. &  \\ 
 & BH$_{sd}$ & $\sqrt{\frac{1}{N-1} \sum_{i=1}^{N} (BH_i - \overline{BH})^2}$ & m & The standard deviation of building heights within the spatial unit. & \\
 & BD & $\frac{\text{building footprint area}}{\text{total land area}}$ & -- & \makecell[l]{Building density (BD) is the proportion of land area covered by \\building footprints.} &  \\
 & FAR & $\frac{\text{total floor area}}{\text{total land area}}$ & -- & \makecell[l]{Floor Area Ratio (FAR) is the ratio of the total floor area \\of all buildings to the total land area.} & \citet{cai2025bi} \\
  & CH & $ \frac{\sum_{i=1}^{n} \text{CH}_i}{n}$ & m & Average tree canopy height in the spatial unit. &  \\
  & CH$_{sd}$ & $\sqrt{\frac{1}{N-1} \sum_{i=1}^{N} (CH_i - \overline{CH})^2}$ & m & The standard deviation of canopy heights within the spatial unit. & \\
  & CD & $\frac{\text{total canopy area}}{\text{total land area}}$ & -- & \makecell[l]{Canopy density (CD) is the proportion of land area covered by canopy.} &  \\

 *2D landscape indices  & PD & $\frac{n_i}{A} \times 10,000$ & patches/100 ha & Patch Density, the density of patches of each landcover class. & \citet{mcgarigal2015fragstats} \\
 & CONTAG & $\left[1 + \sum_{i=1}^{m} \sum_{j=1}^{m} 
\left( \frac{p_{ij} \ln(p_{ij})}{2 \ln m} \right) \right] \times 100$ & $\%$ & \makecell[l]{Contagion index, measuring patch aggregation based on adjacency \\probabilities.} &  \\
 & LSI & $\frac{0.25\sum_{k=1}^{m} e^*_{ik}}{\sqrt{A}}$ & -- & \makecell[l]{Landscape Shape Index measures shape complexity compared to \\a maximally compact shape.} &  \\
 & COHESION & $\left[1 - \frac{\sum_{j=1}^{n} p_{ij}^*}{\sum_{j=1}^{n} p_{ij}^* \sqrt{a^*_{ij}}}\right] \times \left[1 - \frac{1}{\sqrt{Z}}\right]^{-1} \times 100$ & $\%$ & Patch Cohesion Index, cohesion of each landcover class. &  \\
 & PAFRAC & $\displaystyle \frac{n \sum_{i=1}^{n} \ln(p_i)\ln(a_i) - \left( \sum_{i=1}^{n} \ln(p_i) \right)\left( \sum_{i=1}^{n} \ln(a_i) \right)}{n \sum_{i=1}^{n} [\ln(a_i)]^2 - \left( \sum_{i=1}^{n} \ln(a_i) \right)^2}$ & -- & \makecell[l]{Perimeter-Area Fractal Dimension quantifies the complexity of patch \\shapes by examining how patch perimeter scales with patch area.} &  \\
 & CONTIG\_AM & $\frac{\sum_{j=1}^{n} CONTIG_{ij} \cdot a_{ij}}{\sum_{j=1}^{n} a_{ij}}$ & -- & \makecell[l]{Area-Weighted Mean Contiguity Index, measures the spatial \\connectedness of cells within each patch of a landcover class.} &  \\
 & SHDI & $-\sum_{i=1}^{m} p_i \cdot \ln(p_i)$ & -- & \makecell[l]{Shannon Diversity Index, measures the diversity of landcover types\\ based on their proportional abundance.} & \\
 & SHEI & $\displaystyle \frac{-\sum_{i=1}^{m} p_i \ln p_i}{\ln(m)}$ & -- & \makecell[l]{Evenness of class distribution (0 = dominated, 1 = equal).} & \\
 
Environmental variables 
 & DEM & -- & m & Digital Elevation Model (DEM) represents the terrain elevation. & \citet{CopernicusDEM2020} \\
 & Albedo & \makecell[l]{$0.2453\,\rho_{\text{blue}} + 0.0508\,\rho_{\text{green}} + 0.1804\,\rho_{\text{red}} + 0.3081\,\rho_{\text{NIR}}$ \\$+\,0.1332\,\rho_{\text{SWIR1}} + 0.0521\,\rho_{\text{SWIR2}} + 0.0011$} & -- & Ratio of reflected solar radiation to incident solar radiation on a surface. & \citet{cunha2020surface}\\
 & NDVI & $\frac{\rho_{\text{NIR}} - \rho_{\text{red}}}{\rho_{\text{NIR}} + \rho_{\text{red}}}$ & -- & \makecell[l]{Ratio of the difference to the sum of near-infrared and red reflectance, \\used to indicate vegetation vigor and density.} &  \\
 & NDBI & $\frac{\rho_{\text{SWIR}} - \rho_{\text{NIR}}}{\rho_{\text{SWIR}} + \rho_{\text{NIR}}}$ & -- & \makecell[l]{The normalized differential build-up index (NDBI) serves as a \\proxy for the intensity of construction pressure.} & \\
 & WET & \makecell[l]{$0.1511\rho_{\text{blue}} + 0.1972\rho_{\text{green}} + 0.3283\rho_{\text{red}} + $\\$0.3407\rho_{\text{NIR}} - 0.7117\rho_{\text{SWIR1}} - 0.4559\rho_{\text{SWIR2}}$} & -- & \makecell[l]{WET is derived from the Tasseled Cap Transformation, reflecting \\land surface moisture conditions, particularly soil moisture.} & \citet{baig2014derivation} \\
 
 & $D_{sea}$ & $D(x,y) = \min\limits_{(x_w,y_w)\in W} \sqrt{(x-x_w)^2 + (y-y_w)^2}$ & m & Euclidean distance from each pixel to the nearest coastline. & \\

Socio-economic factors & PopD & $\frac{\text{total population}}{\text{total land area}}$ & persons/km$^2$ & Population density & \\
 & RD & $\frac{\text{total road length}}{\text{total land area}}$ & km/km$^2$ & Road density &  
\\
 & IntD & $\frac{\text{ number of intersections}}{\text{total land area}}$ & intersections/km$^2$ & Intersection density &  
\\ 
& RNC & $\frac{\text{ number of intersections}}{\text{total road length}}$ & intersections/km & \makecell[l]{Road network compactness (RNC) measures the density of \\interconnections within the road network.} &  
\\ \hline
\end{tabular}
\begin{tablenotes}
\footnotesize
\item *Note: $a_{ij}$ is the area of patch $j$ of class $i$; 
$A$ is the total landscape area; 
$n_i$ is the total number of patches of class $i$; 
$e^*_{ik}$ is the length of edge between classes $i$ and $k$; 
$p_{ij}^*$ is the perimeter of patch $j$ of class $i$; 
$a_{ij}^*$ is the area of patch $j$ of class $i$; 
$Z$ is the total number of cells in the landscape; 
$c_{ij}$ is the contiguity value of patch $j$ of class $i$; 
$p_i$ is the proportional abundance of class $i$; 
$m$ is the number of classes (patch types).

\end{tablenotes}
\end{threeparttable}
}
\label{tab:varibles}
\end{table}
\end{landscape}

\subsubsection{Land surface temperature retrieval}

Daytime LST is retrieved from Landsat 8 Thermal Infrared Sensor (TIRS) Band 10 using the Statistical Mono-Window algorithm. Prior to LST retrieval, Landsat images are processed using the C Function of Mask (CFMASK) algorithm to remove cloud-contaminated pixels \citep{foga2017cloud}. LST is then estimated using the Statistical Mono-Window Model following \citet{ermida2020google} and \citet{wang2020dynamic}. 

We retrieved LST from Google Earth Engine (GEE), where all required input datasets were assembled and processed. Landsat 8 flies in a sun-synchronous, near-polar orbit with a descending-node equator crossing at around 10:00 a.m. (±15 min) local solar time \citep{usgs_landsat8}. For Singapore (UTC+8; $\sim103.8^{\circ}$E), where civil time runs about +1 h 05 min ahead of local solar time, this corresponds to a typical overpass near 11:15 a.m. Singapore time (SGT). We then aggregated the remaining clear-sky pixels to generate a representative monthly mean LST composite for May from 2018 to 2020 at approximately 11:15 a.m. SGT.

\subsubsection{UTCI calculation}

The Physical-modeled UTCI serves as the humidity-heat stress index and the second target variable in this study, alongside LST. The workflow of $T_{mrt}$ and UTCI calculation is shown in Figure~\ref{fig:utci}. The building height model was derived from a near-official urban building dataset developed by \citet{pei2025parametric}, which is not publicly available due to proprietary restrictions. Tree canopy height was obtained from the 1-m resolution global canopy height maps produced by Meta and the World Resources Institute, which were generated using self-supervised vision transformers and convolutional decoders trained on airborne lidar data \citep{tolan2024canopy}. The dataset was accessed on 10 August 2025 via GEE Community Catalog (\url{https://gee-community-catalog.org/projects/meta_trees/}) and is publicly distributed under the Creative Commons Attribution 4.0 International (CC BY 4.0) license.

SOLWEIG estimates mean radiant flux ($R_{str}$, W\,m$^{-2}$) from six-directional radiation (north, south, east, west, upward, and downward), weighted by the corresponding view factors, surface orientation, and shading, as expressed in Equation~\ref{eq:Rstr}. The resulting total radiant flux is then converted to $T_{mrt}$ 
using the Stefan–Boltzmann relation (Equation~\ref{eq:Tmrt}):

\begin{equation}
R_{\mathrm{str}} = 
\zeta_k \sum_{i=1}^{6} K_i F_i + 
\varepsilon_p \sum_{i=1}^{6} L_i F_i
\label{eq:Rstr}
\end{equation}
\begin{equation}
T_{mrt} = 
\left( 
\frac{R_{\mathrm{str}}}{\varepsilon_p \sigma} 
\right)^{1/4} - 273.15
\label{eq:Tmrt}
\end{equation}

where $K_i$ and $L_i$ denote directional shortwave and longwave radiation fluxes (W\,m$^{-2}$), respectively. $F_i$ are angular view factors. The absorption coefficient for shortwave radiation ($\zeta_k$) was typically 0.7 for a standing person, and the emissivity of the human body ($\varepsilon_p$) was set to 0.97. $\sigma$ is the Stefan–Boltzmann constant ($5.67\times10^{-8}$\,W\,m$^{-2}$\,K$^{-4}$).

Finally, we used the operational UTCI algorithm based on the 6th-order multivariate polynomial published by \citet{brode2012deriving}. Hourly meteorological data at $\approx$2\,km resolution were obtained from the
National Solar Radiation Database (NSRDB) of the National Laboratory of the Rockies (NLR) \citep{sengupta2018national},
including $T_a$, relative humidity, 10-m wind speed, 
global horizontal (GHI), direct normal (DNI), and diffuse horizontal (DHI)
irradiance (Table \ref{tab:nsrdb_variables}). Subsequently, these data were downscaled to 30 m through spatial
interpolation to support city-scale analyses and provide finer inputs for
$T_{\mathrm{mrt}}$ and UTCI mapping. We intentionally avoided generating interpolated high-resolution surfaces from local meteorological station data using approaches such as regression-based data fusion \citep{han2024microclimate,yang2024optimizing}. Such methods rely on auxiliary covariates to reconstruct spatial fields and would therefore embed predictor information (e.g., land use and urban morphology) into the target variable (i.e., UTCI), violating predictor–response independence and confounding the attribution of urban drivers of heat stress. To temporally align with the LST composite, we computed the UTCI using the hourly NSRDB meteorological data. We extracted the specific 11:00 am UTCI maps for every day in May 2019  and averaged them to create a representative monthly mean UTCI map.

\subsubsection{3D urban morphology}
Furthermore, based on the urban building dataset, SVF, floor area ratio (FAR), building density (BD), building height (BH), and standard deviation of building height (BH$_{sd}$), are used to comprehensively characterize 3D building morphology. SVF is a key parameter in urban climatology and environmental studies, especially for high-density urban areas. It is a dimensionless parameter (ranging from 0 to 1) that quantifies the proportion of the sky visible from a given point on the ground or a surface. It reflects the openness of the surrounding environment to the sky and is influenced by urban morphology, such as building height, density, and vegetation. We adopted the grid calculation model following \citet{lindberg2010continuous} to calculate the SVF from buildings and tree canopy. The SVF was computed using the GPU-accelerated algorithm proposed by \citet{li2021gpu}, which implements the grid calculation principles of \citet{lindberg2010continuous} to enable highly efficient, city-scale processing at a 1-m resolution. The ray-casting algorithm was configured with an angular sampling of 360 directions and a search radius of 150 m to accurately capture the shading influence of surrounding urban geometry. The tree canopy was treated as a porous and height-dependent volume. To accurately represent the shading effect of dense tropical foliage, the shortwave transmissivity of light through the tree canopy was set to 3\%, which is consistent with the recommended parameters in the standard SOLWEIG models.

Similarly, tree canopy density (CD), canopy height (CH), including its standard deviation (CH$_{sd}$), are calculated using the tree canopy dataset to comprehensively characterize 3D canopy morphology.

\subsubsection{2D landscape indices}
Drawing on the methodologies established by previous research \citep{chen2024impacts,zhang2024influence,li2024multidisciplinary}, this study selected six widely used landscape indices to assess 2D urban morphology, including percentage of landscape (PLAND), patch density (PD), landscape shape index (LSI), patch cohesion index (COHESION), mean contiguity index (CONTIG), and Shannon diversity index (SHDI), which are calculated based on the land covers that can fully describe urban structural characteristics and spatial configuration of the landscape. The mathematical methods for these calculations are delineated in Table \ref{tab:varibles}. All specified landscape indices were computed using the FRAGSTATS 4.2 software \citep{mcgarigal2002fragstats,mcgarigal2015fragstats}.

\subsubsection{Environmental variables}
Environmental variables, including DEM, albedo, NDVI, normalized difference building index (NDBI), wetness (WET) derived from Landsat 8 Operational Land Imager (OLI), along with the proximity indicator distance to the coastline ($D_{sea}$), are commonly used as indicators for monitoring the surface environment \citep{ramsay2025assessing,tanoori2024machine,xu2024influences}. 

Among them, NDVI is constructed based on the absorption characteristics of vegetation leaves in the red light band and their reflectance in the near-infrared band. It can reflect plant biomass and vegetation coverage. Urbanization, characterized by replacing natural ecosystems with impervious building surfaces, fundamentally reshapes surface energy and moisture fluxes. NDBI is used to delineate built-up and bare-soil features. Similarly, the WET index, derived from the Tasseled Cap Transformation, can reflect surface moisture conditions and evapotranspiration potential, particularly soil moisture \citep{baig2014derivation}.

\subsubsection{Socio-economic factors}
In addition, population density (PopD) was derived from the Resident Population by Planning Area/Subzone of Residence, Ethnic Group and Sex dataset from the Singapore Department of Statistics, based on the Census of Population 2020 \citep{singstat2020population}. While road network density (RD) and road intersection density (IntD) were derived from OpenStreetMap (OSM) data. These variables were used to represent the intensity of human activity \citep{liu2025sensing,liu2026living}.

\subsection{GW-XGBoost model}
\subsubsection{Moran’s I statistic}
Moran’s I statistic (both global and local) was computed as a diagnostic tool to confirm spatial dependence in both target variables and model residuals, thereby justifying the use of spatially explicit machine-learning models. This diagnostic step provides evidence of intrinsic spatial autocorrelation, thereby justifying the use of spatially explicit methods. The value of Moran’s I ranges from $-1$ to 1. A positive Moran’s I indicates positive spatial correlation, and the larger the value, the more pronounced the spatial correlation, while a negative Moran’s I indicates negative spatial correlation. Global Moran’s I index can be calculated as Equation \ref{eq:Moran}.

\begin{equation}
I = \frac{n}{S_0} \cdot \frac{\sum_{i=1}^n \sum_{j=1}^n w_{ij} (x_i - \bar{x})(x_j - \bar{x})}{\sum_{i=1}^n (x_i - \bar{x})^2}
\label{eq:Moran}
\end{equation}
\begin{equation}
S_0 = \sum_{i=1}^n \sum_{j=1}^n w_{ij}
\label{eq:S_0}
\end{equation}

where $n$ denotes the number of spatial units (i.e., subzones), $w_{ij}$ is the spatial weight between locations $i$ and $j$ (determines the degree of influence). $x_i$ is value of the variable at location $i$ and $x_j$ is value of the variable at location $j$. $\bar{x}$ means the global average value of the independent variable at all locations.

Local spatial clusters were identified using the Local Indicators of Spatial Association (LISA) derived from Moran’s I, enabling the identification of spatial clusters (high–high and low–low) and spatial outliers (high–low and low–high) within the study area. A hybrid spatial-weights matrix was adopted, combining Queen contiguity and a nearest-neighbor correction for isolated subzones. Specifically, the weights were primarily based on Queen contiguity, where polygons sharing either a common edge or vertex were considered neighbors. And any isolated polygons (islands) were connected symmetrically to their nearest neighbor based on centroid distance. The final weights matrix was row-standardized prior to analysis.

\subsubsection{Geographically weighted regression (GWR)}

It is essential to consider the spatial heterogeneity of relationships between located variables in the analysis of geographical processes \citep{wang2022gwrboost,yang2023geographically}. GWR is introduced here as a conceptual baseline to motivate geographically weighted machine-learning models, rather than as a primary analytical tool in this study. Unlike traditional regression models that assume relationships between variables are the same across all locations (i.e., spatially stationary relationships), GWR creates a separate regression equation for each specific location. In the classical GWR, a simple linear regression is applied locally to capture the relationship between the dependent (response) variable and the independent (predictor) variables for each location \citep{huang2010geographically}. At each specific location $i$, GWR determines the local regression coefficients by focusing on nearby observations within a defined spatial neighborhood (typically characterized by a bandwidth). These coefficients represent the localized relationship between the dependent variable and the independent variables within the bandwidth area. The general expression for GWR models is given by Equation \ref{eq:gwr}.
\begin{equation}
y_i = \beta_0(u_i, v_i) + \sum_{k=1}^K \beta_k(u_i, v_i) x_{ik} + \epsilon_i
\label{eq:gwr}
\end{equation}

where $y_i$ represents the dependent variable for the $i^{th}$ location, $(u_i, v_i)$ denotes the coordinates of the $i^{th}$ location, $\beta_0$ is the constant term, $\beta_k(u_i, v_i)$ ($k = 1, \ldots, K$) represents the local regression coefficients of the independent variables at the location $i$, $K$ is the total number of independent variables (excluding the intercept), and $\epsilon_i$ is the random residual which usually obeys a normal distribution (Gaussian distribution) hypothesis. The observations in the neighborhood location $\beta(u_j, v_j)$ of location $i$ are collected to estimate the parameters $\beta(u_i, v_i)$ in a geographically weighted least square approach (Equation \ref{eq:gwls}):
\begin{equation}
\hat{\beta}(u_i, v_i) = \left(\bm{X}^T \bm{W}(u_i, v_i) \bm{X}\right)^{-1} \bm{X}^T \bm{W}(u_i, v_i) \bm{y}
\label{eq:gwls}
\end{equation}
\begin{equation}
\bm{X} =
\begin{bmatrix}
1 & x_{11} & x_{12} & \cdots & x_{1k} \\
1 & x_{21} & x_{22} & \cdots & x_{2k} \\
\vdots & \vdots & \vdots & \ddots & \vdots \\
1 & x_{n1} & x_{n2} & \cdots & x_{nk}
\end{bmatrix}
\label{eq:X}
\end{equation}
\begin{equation}
\bm{W}(u_i, v_i) =
\begin{bmatrix}
w_1(u_i, v_i) & 0 & 0 & \cdots & 0 \\
0 & w_2(u_i, v_i) & 0 & \cdots & 0 \\
0 & 0 & w_3(u_i, v_i) & \cdots & 0 \\
\vdots & \vdots & \vdots & \ddots & \vdots \\
0 & 0 & 0 & \cdots & w_n(u_i, v_i)
\end{bmatrix}
\label{eq:W}
\end{equation}

where $\hat{\beta}(u_i, v_i)$ denotes the estimate of the location-specific parameter, $\bm{X}$ (Equation \ref{eq:X}) and $\bm{y}$ are matrixes of independent variables and dependent variable for neighboring study area determined by the kernel bandwidth, respectively, n is the number of observations (rows), and the matrix has dimensions n × (k + 1) because it includes the intercept column.
$\bm{W}(u_i, v_i) \in \mathbb{R}^{N \times N}$ in Equation \ref{eq:W} indicates the diagonal spatial weights matrix generated by kernel functions for total $N$ observations. Generally, the off-diagonal elements of $\bm{W}(u_i, v_i)$ are zero, and the diagonal elements of $\bm{W}(u_i, v_i)$ denote the geographical weight of $N$ samples for the specific observation $i$. The coefficients of the same independent variable vary across the study area because of the respective estimates at each sample location. And such coefficients can be used in further spatial analysis.

Besides, selecting an appropriate kernel bandwidth is crucial in GWR, as the model’s performance is highly sensitive to this parameter \citep{li2010investigating}. In this study, the optimum bandwidth (or the best neighbor size) is determined by minimizing the Akaike Information Criterion (AIC) values. A bi-square kernel with an adaptive distance approach was employed to capture microclimatic variations, as it allows the spatial influence of neighboring observations to vary with local data density in compact and highly heterogeneous urban environments such as Singapore. The weight using the bi-square kernel function is expressed as Equation \ref{eq:weight}:

\begin{equation}
w_j(u_i, v_i) =
\begin{cases}
\left(1 - \left(\frac{d_{ij}}{b}\right)^2\right)^2, & \text{if } d_{ij} \leq b, \\
0, & \text{if } d_{ij} > b,
\end{cases}
\label{eq:weight}
\end{equation}

where \(w_j(u_i, v_i)\) is spatial weight for observation \(j\) at the target location \(u_i, v_i\), \(d_{ij}\) is the distance between observation \(j\)and the target location\(i\), \(b\) is the bandwidth parameter, defining the maximum distance where weights are non-zero and controlling the spatial extent of the weights.

\subsubsection{XGBoost global model}
As proposed by \citet{chen2016xgboost}, the XGBoost algorithm is a scalable and efficient tree boosting framework designed to address the limitations of traditional gradient boosting methods. It enhances prediction accuracy by iteratively building Classification and Regression Trees (CARTs) that address the residual errors of previously constructed trees. The foundation of the XGBoost algorithm is the concept of boosting, an ensemble learning approach that aims to integrate multiple weak learners to form a strong learner. Each weak learner in the ensemble is sequentially trained to correct the errors made by its predecessors. In other words, each new CART is trained to fit the residuals of the preceding CARTs, thereby incrementally reducing the overall error. For a given set of predictions \(\hat{y}\), boosting minimizes the following objective as Equation \ref{eq:xgboost} and \ref{eq:xgboost2}:

\begin{equation}
\mathcal{L}^{(t)} =
\sum_{i=1}^{n}
\ell\!\left(
y_i,\,
\hat{y}_i^{(t-1)} + f_t(x_i)
\right)
+ \Omega(f_t),
\label{eq:xgboost}
\end{equation}

\begin{equation}
\hat{y}_i = \sum_{t=1}^T f_t(x_i)
\label{eq:xgboost2}
\end{equation}

where \(\ell\) is the loss function that quantifies the difference between the observed (\(y_i\)) and predicted (\(\hat{y}_i^{(t)}\)) values for the $i^{th}$ observation, \(f_t(x_i)\) is the output of the \(t^{th}\) tree for the \(i^{th}\) observation, \(T\) is the total number of trees in the ensemble, 
and \(\Omega(f_t)\) is the regularization term of the \(t^{th}\) tree that penalizes model complexity to prevent overfitting.

\subsubsection{GW-XGBoost model}
The GW-XGBoost, which combines the concepts of GWR and XGBoost, is employed for analysis due to its potential to address the limitations of the GWR model and improve predictive performance over a non-geographically weighted XGBoost model. Similar to the regression analysis framework of GWR, GW-XGBoost consists of multiple sub-models calibrated locally using XGBoost instead of linear regression as shown in Equation \ref{eq:gw-xgboost} and \ref{eq:gw-xgboost2}.

\begin{equation}
\mathcal{L}^{(t)}(u_i, v_i)
=
\sum_{j=1}^{n}
w_j(u_i, v_i)\,
\ell\!\left(
y_j,\,
\hat{y}_j^{(t-1)} + f_t(x_j)
\right)
+ \Omega(f_t)
\label{eq:gw-xgboost}
\end{equation}

\begin{equation}
\hat{y}_j = \sum_{t=1}^{T} f_t(x_j)
\label{eq:gw-xgboost2}
\end{equation}

where \(w_j(u_i, v_i)\) is the spatial weight for observation \(j\) relative to location \((u_i, v_i)\), computed as Equation \ref{eq:weight}. Although Equations \ref{eq:xgboost2} and \ref{eq:gw-xgboost2} share the same additive prediction form, the regression trees \(f_t\) in Equations \ref{eq:gw-xgboost2} are estimated locally using spatially weighted loss functions, and therefore differ across target locations. Moreover, in GW-XGBoost, the intercept is not explicitly modeled, as decision trees inherently account for offsets through recursive partitioning and leaf values. Consequently, an explicit intercept term is unnecessary, and the first column of \(\bm{X}\)in Equation \ref{eq:X} can be removed. The resulting feature matrix \(\bm{X}_{\text{modified}}\) is defined in Equation \ref{eq:X_mod}.

\begin{equation}
\bm{X}_{\text{modified}} =
\begin{bmatrix}
x_{11} & x_{12} & \cdots & x_{1k} \\
x_{21} & x_{22} & \cdots & x_{2k} \\
\vdots & \vdots & \ddots & \vdots \\
x_{n1} & x_{n2} & \cdots & x_{nk}
\end{bmatrix}.
\label{eq:X_mod}
\end{equation}

For each target location, GW-XGBoost fits a local XGBoost model by emphasizing nearby observations through distance-based sample weights, while retaining the nonlinear learning capacity and interpretability of gradient-boosted trees. Specifically, spatial dependence is incorporated through instance-weighted loss minimization, where observations closer to the target location \((u_i, v_i)\) receive stronger influence during model training. Specifically, spatial weights \(w_j(u_i, v_i)\) were passed as sample weights to the local XGBoost training procedure, which natively supports instance weighting. This approach preserves the integrity of tree-based split criteria while allowing geographically proximate observations to contribute more strongly to local model calibration.

For each target location, GW-XGBoost fits a local XGBoost model using the original feature matrix \(\bm{X}_\text{modified}\), with spatial proximity encoded solely through sample weights rather than feature re-scaling. In this way, the framework retains the nonlinear learning capacity of gradient-boosted trees while explicitly accounting for spatial heterogeneity.

To provide an overview of the pairwise relationships among LST, UTCI, and all urban factors used in this study, we included a Pearson correlation matrix (Figure \ref{fig:Pearson}). This figure helps illustrate the degree of association among predictors and provides additional context for interpreting the modeling results. Model interpretability in the GW-XGBoost framework is achieved through SHAP, which quantifies each local prediction into additive feature contributions expressed in consistent units of the response variable, thereby enabling spatially explicit interpretation of model behavior. Details of the SHAP-based analysis are presented in Section~3.3.

\subsubsection{Model evaluation}
To ensure robust model evaluation and avoid overfitting during hyperparameter tuning, a nested cross-validation (Nested CV) procedure was firstly employed for the global XGBoost model \citep{grekousis2025geographical}. The framework consists of two loops: an inner loop that performs hyperparameter optimization and an outer loop that evaluates model generalization on held-out subsets. In the inner loop, combinations of three primary XGBoost hyperparameters, i.e., the number of trees ($n\_estimators$), learning rate ($learning\_rate$), and maximum tree depth ($max\_depth$), were systematically tested using grid search. The configuration yielding the lowest root-mean-square error (RMSE) within each training subset was selected. In the outer loop, the optimized hyperparameters obtained from the inner cross-validation were evaluated across five independent test folds to assess the model’s generalization performance. 
This hierarchical procedure provides an unbiased estimate of the model’s predictive capability by ensuring that hyperparameter selection and model validation are conducted on separate data partitions. The resulting performance metrics (R$^2$, MAE, and RMSE) were averaged across outer folds to quantify both model accuracy and generalization stability. 
After completing the nested CV procedure, the model was retrained using the optimal hyperparameters on the entire training dataset and subsequently evaluated on an independent hold-out test set (not involved in any CV step). These globally optimized hyperparameters were then applied uniformly across the GW-XGBoost models to ensure cross-location comparability; however, because the present implementation did not impose strict block-based spatial CV, some residual optimism in predictive accuracy due to spatial autocorrelation may remain. The hyperparameter settings for XGBoost and GW-XGBoost are shown in Table \ref{tab:xgboost_hyperparameters}.

The local GW-XGBoost model was validated through spatial cross-validation during the bandwidth optimization process. Tree-level hyperparameters were inherited from the global nested CV search to prevent over-tuning at the local level, while the local component was governed by the bandwidth ($bw$) and kernel weighting. An adaptive bisquare kernel (as shown in Equation \ref{eq:weight}) was employed to define local neighborhoods, and the optimal bw was determined by minimizing cross-validation loss using a leave-one-out spatial CV \citep{wong2015performance,pang2023enhanced,cawley2008efficient}. For each candidate $bw$, the model predicted each observation using all other locations, with distance-based weights applied to nearby samples. The final model was trained with the optimal $bw$ and used to generate diagnostic maps of local $R^2$ and standardized residuals to evaluate spatial heterogeneity in model performance. To rigorously evaluate the predictive performance of the GW-XGBoost models while preventing overfitting, we utilized a pseudo-out-of-bag validation approach inherent to Stochastic Gradient Boosting. Unlike standard Random Forests, which utilize bootstrap aggregating (bagging), standard gradient boosted trees typically process all training data sequentially. To generate an internal validation metric, the subsample hyperparameter was set to 0.8. Consequently, during the construction of each localized model—which is trained using data from the target subzone and its 94 nearest neighbors defined by the adaptive kernel—the algorithm stochastically samples exactly 80\% of the available local instances to grow each individual decision tree. The remaining 20\% of the instances are temporarily withheld as the test set for that specific iteration. As the boosting sequence progresses, the model continuously predicts the target variable for these stochastically withheld, unseen instances.

\section{Results}
\subsection{Spatial distribution of LST and UTCI}

\begin{figure}[h!]
    \centering   \includegraphics[width=1.0\linewidth]{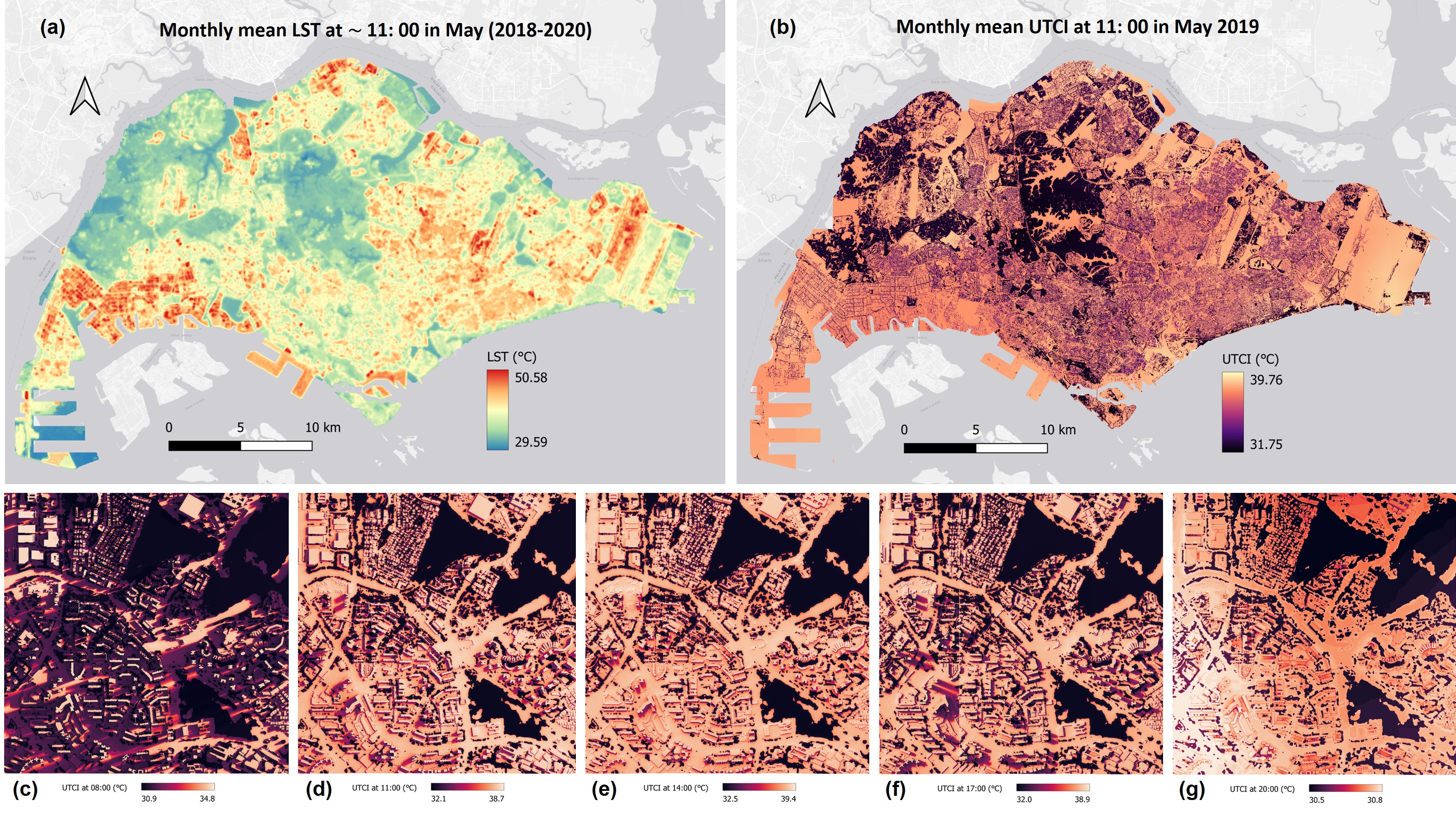}
    \caption{Spatial distribution of 30-m LST (a) and 1-m UTCI (b), and the dynamic variations of UTCI at 8:00, 11:00, 14:00, 17:00, and 20:00 in a zoomed-in patch area (c-g).}
    \label{fig:LST_UTCI}
\end{figure}

Figure~\ref{fig:LST_UTCI} compares the spatial patterns of 30-m LST (Figure~\ref{fig:LST_UTCI}a) and 1-m UTCI (Figure~\ref{fig:LST_UTCI}b) at the city scale across the main island of Singapore, highlighting substantial differences between surface radiative temperatures and human heat stress. Overall, LST retrieved from Landsat 8 TIRS demonstrates both higher maximum (with temperatures exceeding 50\,$^\circ$C) and lower minimum values than UTCI. The LST map reveals pronounced surface overheating in industrial and residential areas within industrial and residential clusters in the western, eastern, and northern regions. In contrast, UTCI represents a more moderate spatial thermal condition experienced by humans at the pedestrian level. The 1-m UTCI map shows finer variations influenced by shading, vegetation, and local morphology. The UTCI values ranging from 31.75\,$^\circ$C to 39.76\,$^\circ$C predominantly fall within the categories of \textit{strong heat stress} (32--38\,$^\circ$C) and \textit{very strong heat stress} (38--46\,$^\circ$C) \citep{brode2012deriving}.
The zoomed-in panels in Figure \ref{fig:LST_UTCI}c–g further present the diurnal dynamics of UTCI in a patch area of the Clementi neighborhood, capturing the temporal variations at 08:00, 11:00, 14:00, 17:00, and 20:00. The coverage and direction of shading are changing with time. Higher UTCI values are observed around midday (14:00) in open and impervious areas, consistent with findings reported by \citet{park2014application}, while cooling effects from tree canopy and building shadows are more evident in the early morning and late afternoon.

\begin{figure}[h!]
    \centering   \includegraphics[width=0.8\linewidth]{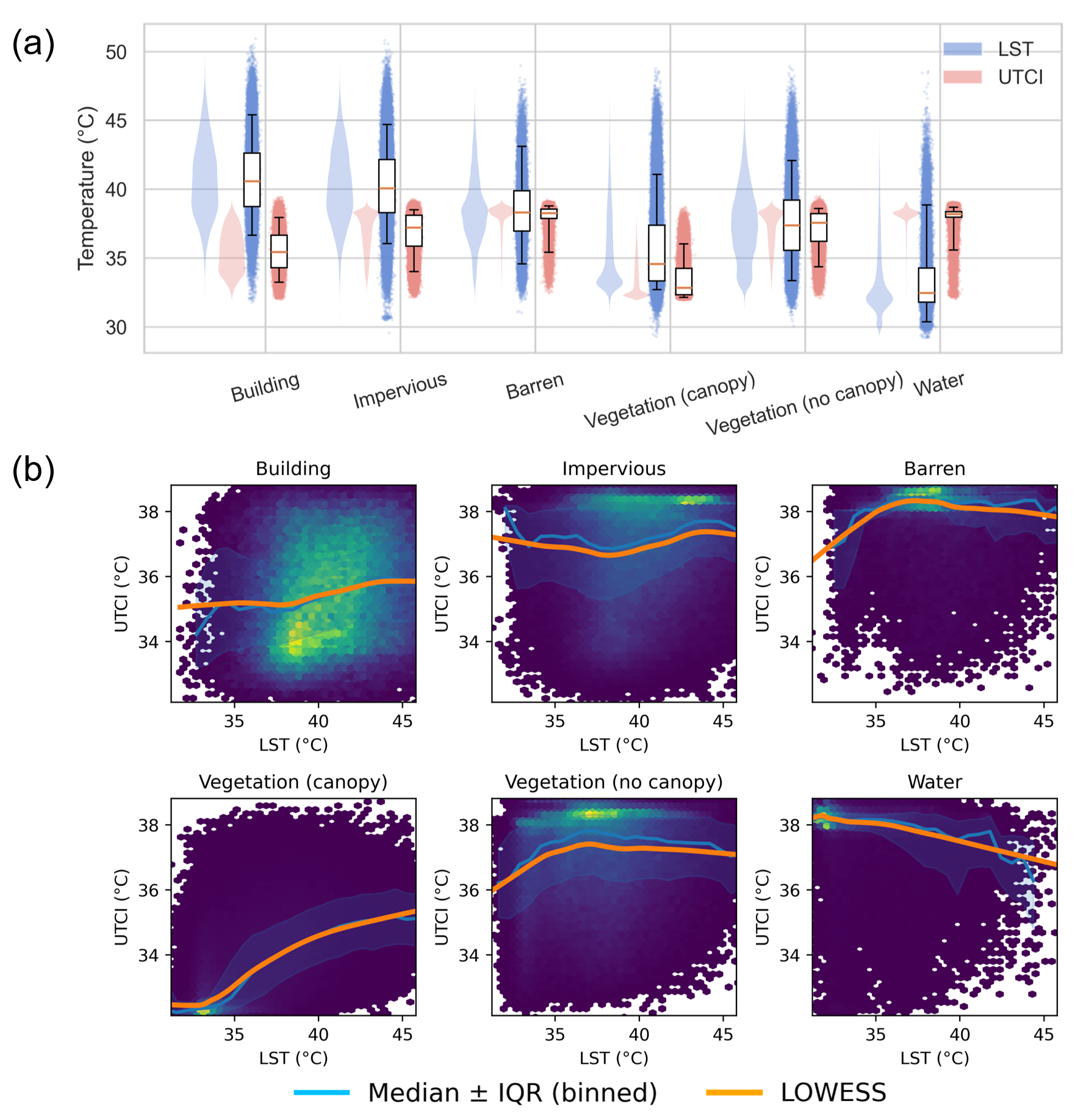}
    \caption{Comparison between LST and human-centric UTCI across LULC types.}    
    \label{fig:LST_UTCI_LC}
\end{figure}

Figure~\ref{fig:LST_UTCI_LC} compares LST and human-centric heat stress across land use/land cover (LULC) types. The double violins and box plots in Figure~\ref{fig:LST_UTCI_LC}a show the probability density, median and quartile values of LST and UTCI for each LULC class at 30 m resolution. Built-up and impervious areas show the highest LST, indicating strong surface heating effects. However, their corresponding UTCI values are relatively lower, particularly in build-up areas. Vegetated areas with canopy cover show markedly lower LST and UTCI values than other LULC types, and the consistently lower UTCI further highlights the cooling influence of shading and evapotranspiration. Moreover, barren surface and vegetation without canopy show similar LST and UTCI magnitudes. In contrast, water bodies exhibit the lowest LST but much higher and stable UTCI, reflecting the decoupling between surface temperature and near-surface thermal conditions. 

Similarly, Figure~\ref{fig:LST_UTCI_LC}b compares two nonparametric visualizations of the LST-UTCI relationship across LULC types. The binned median with the interquartile range (IQR, 25th–75th percentile) highlights the variability of UTCI within each LST bin, while the locally weighted scatterplot smoothing (LOWESS) regression reveals the smooth overall nonlinear trends \citep{cleveland1979robust,moran1984locally}. Both show that UTCI generally increases with LST, but the relationship is non-linear and saturates at higher LST values. However, water shows an opposite and negative correlation, which is most prominent during hot conditions. Furthermore, buildings and impervious surfaces exhibit large scatter and wide IQRs, providing evidence of microclimatic heterogeneity due to shading and material differences.

\begin{figure}[h!]
    \centering   \includegraphics[width=1.0\linewidth]{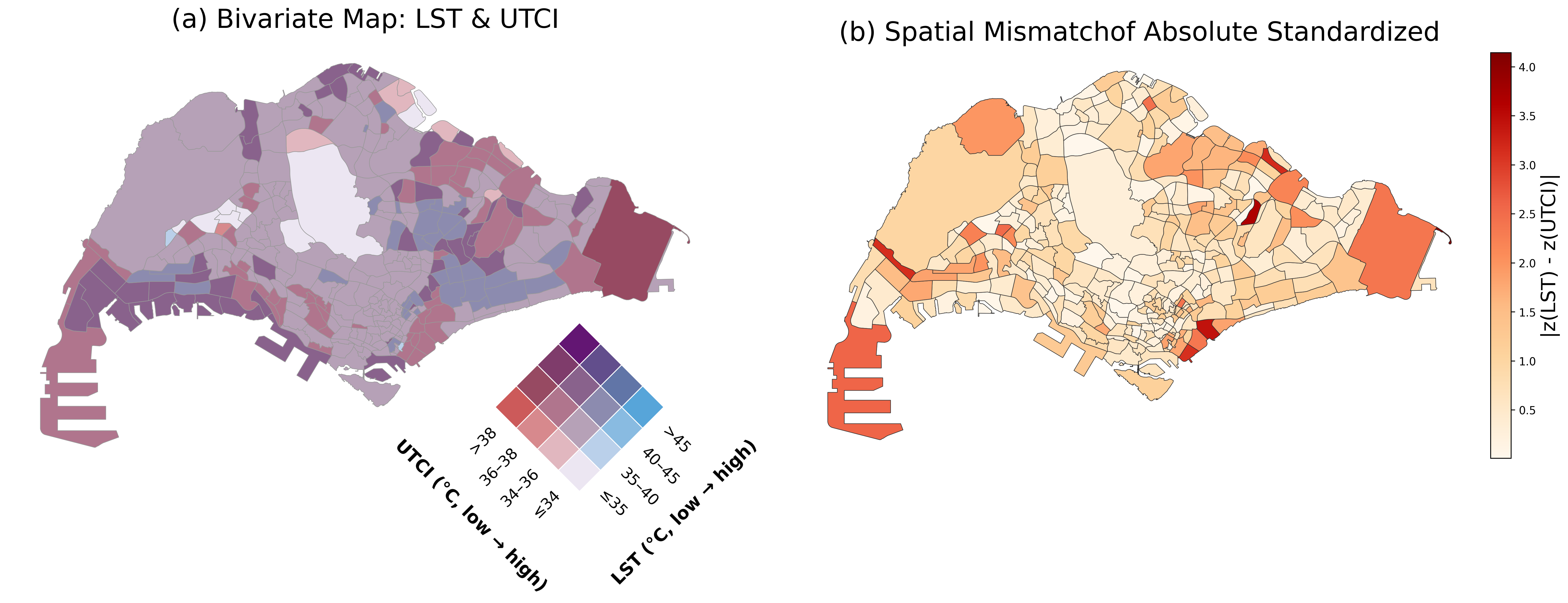}   \caption{Spatial relationship and quantitative divergence between surface temperatures and human-centric heat stress. (a) Bivariate choropleth map illustrating the coupled spatial distribution of LST and UTCI. The rotated diamond legend categorizes intersecting thermal profiles into four primary extremes: (1) Bottom corner (white): Low LST and low UTCI, representing cool and comfortable environments; (2) Top corner (dark purple): High LST and high UTCI, indicating severe, compounding thermal risk; (3) Right corner (blue): High LST but low UTCI, representing hot physical surfaces where human heat stress is successfully mitigated (e.g., via high ventilation); and (4) Left corner (red/pink): Low LST but high UTCI, representing cool surfaces that nonetheless experience severe pedestrian heat stress (e.g., due to trapped radiation or lack of shade). (b) Spatial distribution of the absolute standardized mismatch per subzone ($|z(\text{LST}) - z(\text{UTCI})|$). Dark orange areas highlight severe quantitative discrepancies.}    \label{fig:bivariate_LST_UTCI}
\end{figure}

Water bodies and rooftop pixels were retained in the pixel-level LULC comparison shown in Figure~\ref{fig:LST_UTCI_LC}, but were masked prior to aggregation for the subzone-level analysis of LST, UTCI, and all environmental variables used in GW-XGBoost. This ensured that the modeling dataset represented terrestrial, pedestrian-accessible environments relevant to human heat exposure assessment. Figure~\ref{fig:bivariate_LST_UTCI} presents a side-by-side comparison of the categorical thermal risk and the quantitative mismatch. Figure~\ref{fig:bivariate_LST_UTCI}a utilizes a bivariate choropleth map to illustrate the combined spatial distribution of LST and UTCI, while Figure~\ref{fig:bivariate_LST_UTCI}b maps the absolute standardized difference ($|z(\text{LST}) - z(\text{UTCI})|$). The spatial distribution in Figure~\ref{fig:bivariate_LST_UTCI}b shows that the most severe discrepancies (dark orange regions) are not randomly distributed but are concentrated in specific urban areas that exhibit relatively low LST yet elevated UTCI.

\subsection{The performance of spatial machine learning models}

\begin{figure}[h!]
    \centering   \includegraphics[width=0.8\linewidth]{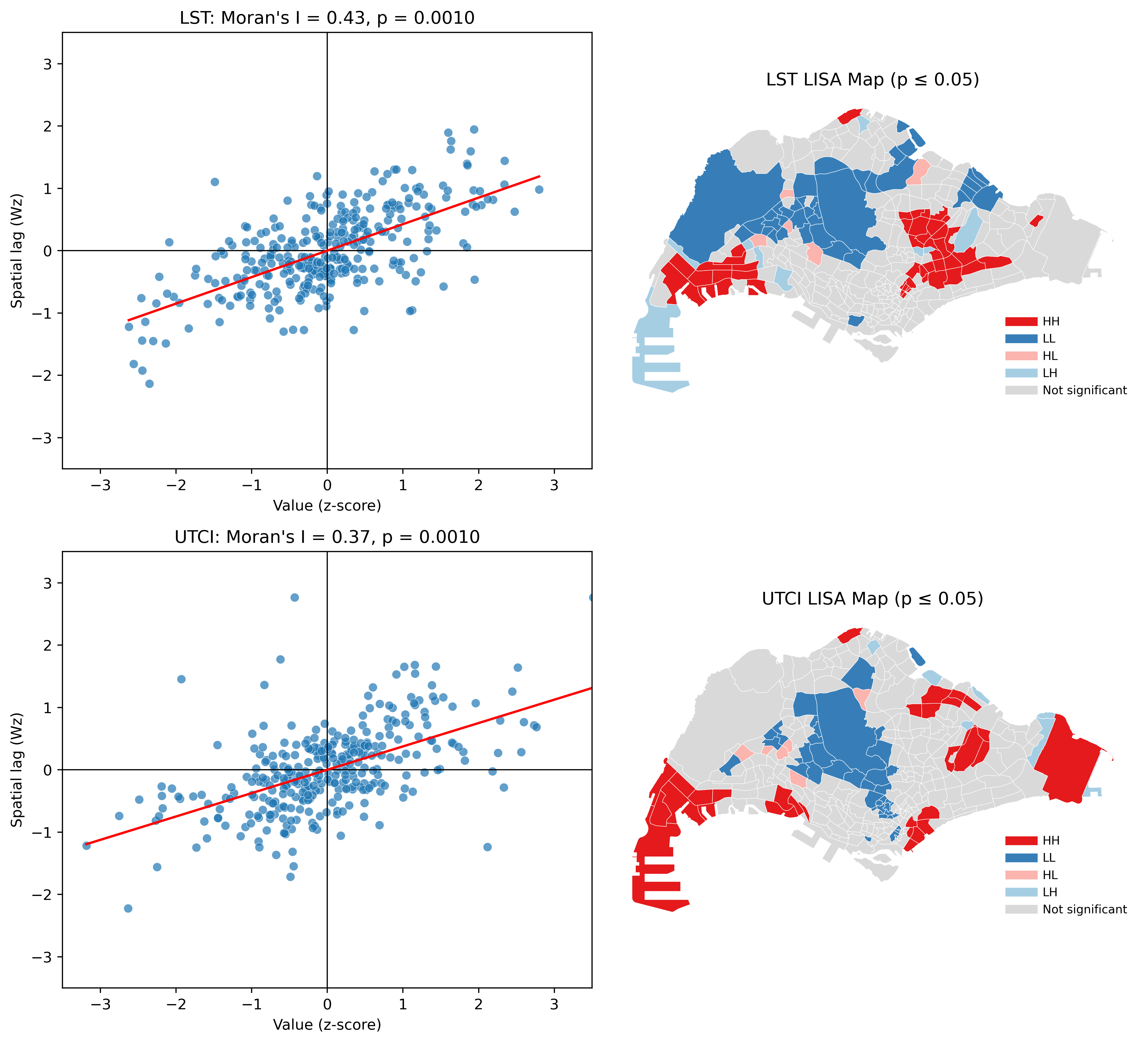}
    \caption{Global and local Moran’s I statistics for LST and UTCI. A hybrid Queen–nearest-neighbor spatial-weights matrix is used, where isolated subzones are symmetrically connected to their nearest polygon to ensure full spatial connectivity.}
    \label{fig:mi}
\end{figure}

Figure~\ref{fig:mi} presents the spatial autocorrelation of LST (upper row) and UTCI (lower row) mean values across subzones in Singapore, using both Moran’s I scatter plots and Local Indicators of Spatial Association (LISA) cluster maps. Both global and local spatial autocorrelation analyses reveal that LST and UTCI exhibit significant positive spatial dependence across subzones (Moran’s I = 0.43 and 0.37, $p < 0.05$), with high LST clusters primarily concentrated in eastern dense urban residential and western industrial areas and low LST clusters in vegetated and coastal areas, while high UTCI is mainly distributed in southern city center, eastern resident area, western industrial areas, and port/airport.

Table \ref{tab:performance} compares the predictive performance of the global XGBoost model and the GW-XGBoost for estimating LST and UTCI. For both targets, the XGBoost models exhibited strong predictive performance, achieving test set $R^2$ values of 0.872 for LST and 0.831 for UTCI. The corresponding MAE of 0.563~$^\circ$C and 0.188~$^\circ$C, respectively, indicate small prediction deviations. The GW-XGBoost model for LST yielded comparable aggregated predictive performance (test $R^2 = 0.855$, MAE = 0.600~$^\circ$C, RMSE = 0.808~$^\circ$C), showing only a modest difference from global XGBoost, whereas enhancing local interpretability by capturing spatially varying relationships with a mean local $R^2$ of 0.750~$\pm$~0.108. For UTCI, GW-XGBoost improved fit precision (test $R^2 = 0.905$, MAE = 0.172~$^\circ$C, RMSE = 0.261~$^\circ$C) and exhibited a mean local $R^2$ of 0.858~$\pm$~0.057, highlighting its superior ability to represent spatial heterogeneity. To assess spatial autocorrelation, we calculated the Global Moran’s I for the residuals of both Global XGBoost and GW-XGBoost (Figure \ref{fig:model_residuals}). For LST, the residual autocorrelation was weak and non-significant in both models (Moran’s I = 0.012, $p = 0.306$ and I = 0.034, $p = 0.157$, respectively), indicating both models already showed minimal residual spatial dependence for LST. For UTCI, the Global XGBoost residuals showed a marginal tendency toward clustering ($I$ = 0.037, $p = 0.087$). However, the implementation of the GW-XGBoost model more effectively accounted for the remaining spatial heterogeneity in UTCI ($I$ = -0.015, $p = 0.372$). Overall, although the global models performed well, the GW-XGBoost framework maintained rigorous spatial randomness in the residuals while also enabling the identification of highly localized and spatially varying feature relationships.

\begin{table*}[htbp]
\centering
\caption{Comparison of model performance between XGBoost and geographically weighted XGBoost (GW-XGBoost) for predicting LST and UTCI.}
\label{tab:performance}
\resizebox{\linewidth}{!}{
\begin{threeparttable}
\renewcommand{\arraystretch}{1.20}
\begin{tabular}{lllllllll}
\toprule
\textbf{Target} & \textbf{Model} & \textbf{Evaluation} & \textbf{$R^2$} & \textbf{MAE (°C)} & \textbf{RMSE (°C)} & \textbf{Main Hyperparameters} & \textbf{Kernel} & \textbf{Bandwidth} \\
\midrule
\multirow{5}{*}{\textbf{LST}} 
 & \multirow{3}{*}{\textbf{XGBoost}} 
 & Nested CV (mean ± SD) & 0.859 ± 0.032 & 0.582 ± 0.073 & 0.818 ± 0.051 & 
   \multirow{3}{*}{\begin{tabular}[c]{@{}l@{}}$n_{\text{estimators}}$ = 500;\\ learning\_rate = 0.05;\\ max\_depth = 2\end{tabular}} & – & – \\
 &  & Test set & \textbf{0.872} & \textbf{0.563} & \textbf{0.757} &  & – & – \\
 & \multirow{2}{*}{\textbf{GW-XGBoost}} 
 & Mean local goodness-of-fit & 0.750 ± 0.108 & 0.022 ± 0.021 & 0.817 ± 0.091 & 
   \multirow{2}{*}{} & Adaptive & 94 (best) \\
 &  & Test set & \textbf{0.855} & \textbf{0.600} & \textbf{0.808} &  &  &  \\
\multirow{5}{*}{\textbf{UTCI}} 
 & \multirow{3}{*}{\textbf{XGBoost}} 
 & Nested CV (mean ± SD) & 0.898 ± 0.049 & 0.176 ± 0.006 & 0.273 ± 0.022 & 
   \multirow{3}{*}{\begin{tabular}[c]{@{}l@{}}$n_{\text{estimators}}$ = 500;\\ learning\_rate = 0.05;\\ max\_depth = 2\end{tabular}} & – & – \\
 &  & Test set & \textbf{0.831} & \textbf{0.188} & \textbf{0.336} &  & – & – \\
 & \multirow{2}{*}{\textbf{GW-XGBoost}} 
 & Mean local goodness-of-fit & 0.858 ± 0.057 & 0.007 ± 0.006 & 0.261 ± 0.038 & 
   \multirow{2}{*}{} & Adaptive & 94 (best) \\
 &  & Test set & \textbf{0.905} & \textbf{0.172} & \textbf{0.261} &  &  &  \\
\bottomrule
\end{tabular}%
\begin{tablenotes}
\footnotesize
\item *Note:CV = cross-validation; Mean ± SD values indicate model stability across folds (for global) or across subzones (for local). All models used identical hyperparameter settings for comparability. The adaptive kernel bandwidth of 94 was selected as optimal.
\end{tablenotes}
\end{threeparttable}
}
\end{table*}

\begin{figure}[h!]
    \centering       \includegraphics[width=1.0\linewidth]{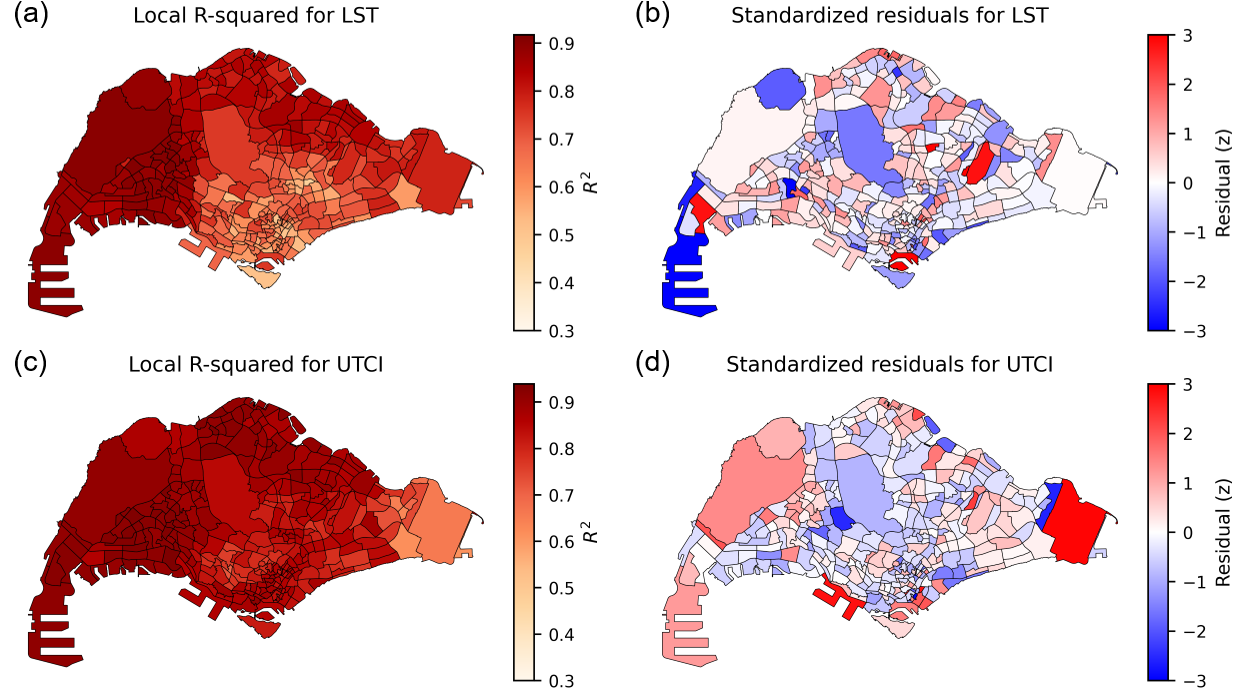}
    \caption{Spatially local R$^2$ and standardized residuals of the GW-XGBoost model for LST (a-b) and UTCI (c-d) across subzones of Singapore.}
    \label{fig:GWXGB_r2}
\end{figure}

Figure~\ref{fig:GWXGB_r2} visualizes the spatially explicit $R^2$ and standardized residuals of the GW-XGBoost models for LST and UTCI, respectively. Both models achieved high local explanatory power (local $R^2 >$ 0.64 for LST and local $R^2 >$ 0.80 for UTCI, respectively) across planning subzones, with slightly lower performance in the central region for LST and in the airport for UTCI. Standardized residuals also show localized over-predictions LST in the western Tuas Port and forest area (Figure~\ref{fig:GWXGB_r2}b), while under-predictions UTCI in the airport, Pasir Panjang Terminals Port, and the western water catchment with many impervious and barren lands and no-canopy vegetation that lack shading (Figure~\ref{fig:GWXGB_r2}d).

Overall, the GW-XGBoost results indicate that UTCI exhibits a systematically stronger sensitivity to urban spatial structure than LST. While LST spatial variation is largely driven by land cover types \citep{wang2024integrated}, the superior predictive accuracy of our spatial model for UTCI suggests that human heat stress is intricately governed by 3D morphological arrangements. This heightened structural sensitivity underscores the potential for targeted urban design interventions to mitigate heat stress, a theme further explored through feature importance analysis in Section \ref{shap_importance}.

\subsection{The difference of influencing factors in LST and UTCI revealed by explainable spatial machine learning}
\label{shap_importance}

\begin{figure}[h!]
    \centering       \includegraphics[width=1.0\linewidth]{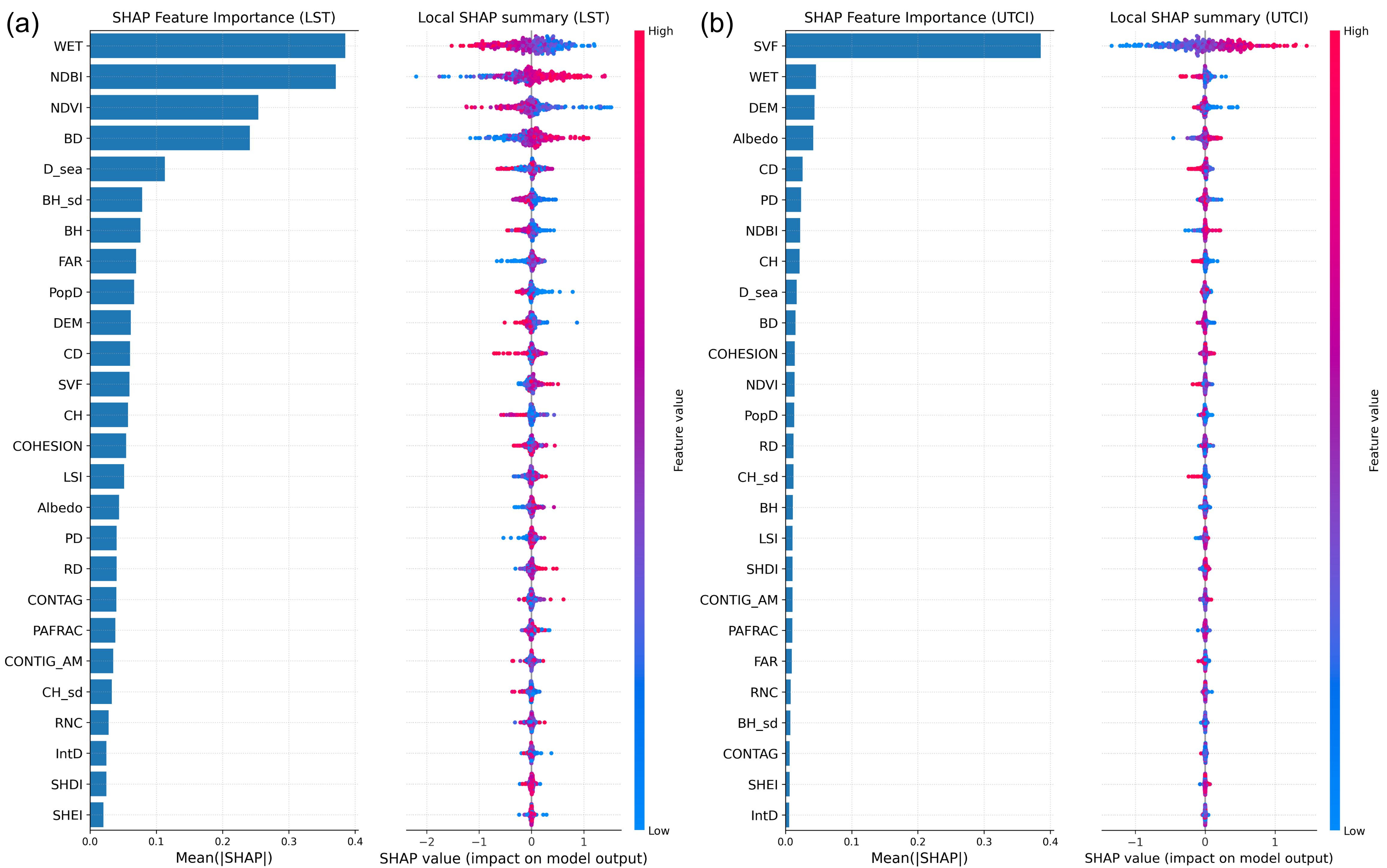}
    \caption{SHAP global feature importance and local summary (beeswarm) plots for LST (a) and UTCI (b).}
    \label{fig:GWXGB_SHAP}
\end{figure}

\begin{figure}[h!]
    \centering       \includegraphics[width=1.0\linewidth]{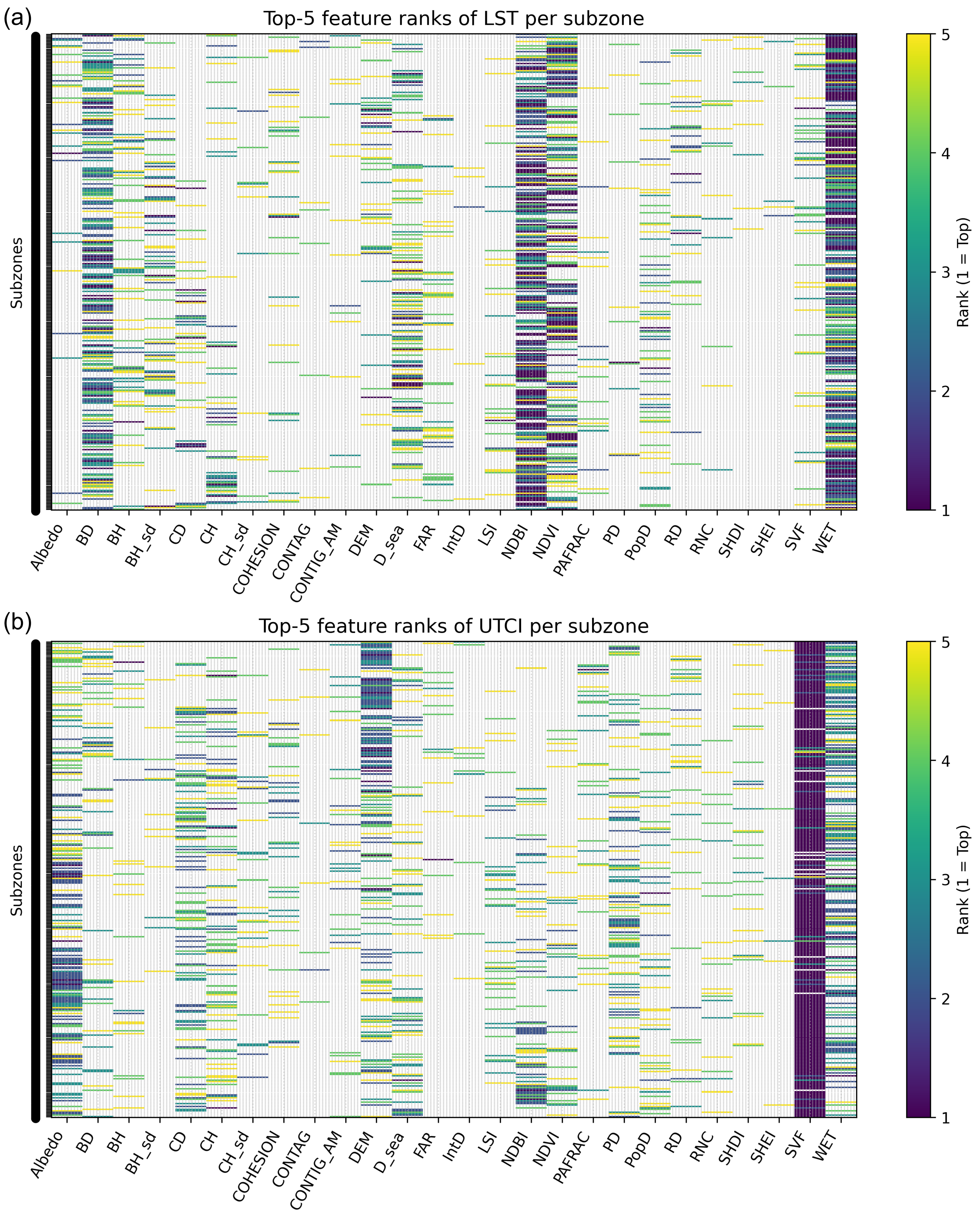}
    \caption{Spatial variation of the top-five feature importance ranks across subzones for (a) LST and (b) UTCI. Each column corresponds to a predictor, and each row to a subzone. Colors represent the local ranking of each feature (1 = Top).}
    \label{fig:top_feature}
\end{figure}

The global feature importance and local summaries derived from the SHAP framework are visualized in Fig~\ref{fig:GWXGB_SHAP}. For LST, environmental variables such as WET, NDBI, and NDVI dominate the predictions, highlighting the trade-off between vegetation and imperviousness. Whereas, morphological variables (SVF and DEM), radiative factor (Albedo), and WET are more influential for UTCI, emphasizing the role of shading and vertical openness in modulating human physiological heat exposure. Figure \ref{fig:top_feature} further compares the Top-5 ranked features influencing LST and UTCI for each subzone. The ranking patterns vary considerably across subzones, indicating that the spatial heterogeneity in dominant predictors of surface and thermal comfort temperatures.

Moreover, Figure~\ref{fig:GWXGB_features} compares local variable dominance and SHAP-based effects for GW-XGBoost models of LST and UTCI in each subzone. Figure~\ref{fig:GWXGB_features}a shows the conventional tree-based primary gain importance variables in each subzone, indicating the model's performance relied on them most to explain spatial temperature variation, where five factors out of 26 predictor variables exert the strongest local influence on improving the local R$^2$ of the LST model across all 328 subzones. However, 3D urban morphological variables are not among the primary variables of importance, indicating a limitation of LST in reflecting 3D urban structure. In contrast, Figure~\ref{fig:GWXGB_features}d shows that 3D metrics, including SVF, canopy height, and the standard deviation of canopy height, have the most important influence on a higher R$^2$ of UTCI and cover most built-up areas.

\begin{figure}[h!]
    \centering    \includegraphics[width=1.0\linewidth]{Figure10_subzone.png}
    \caption{Spatially explicit results of the GW-XGBoost model for LST and UTCI across subzones of Singapore. (a-c) Primary gain, Primary SHAP, and SHAP value of SVF contributing to LST in each subzone; (d-f) Primary gain, Primary SHAP, and SHAP value of SVF contributing to UTCI in each subzone. All these maps illustrate spatial heterogeneity in the controlling factors of LST and UTCI and the reliability of local model fits.}
    \label{fig:GWXGB_features}
\end{figure}

Furthermore, Figure~\ref{fig:GWXGB_features}(b,e) measures actual contribution magnitude (mean $|\text{SHAP}|$) of each predictor per subzone and quantifies the marginal effect of each predictor on the target variable in consistent units of the response variable (e.g.,~$^\circ$C). Figure~\ref{fig:GWXGB_features}b shows that the primary SHAP-based importance features are mainly WET, NDVI, NDBI, and BD. Furthermore, Figure~\ref{fig:GWXGB_features}e confirms SVF as the top explanatory variable across most subzones. 

Moreover, the signed SHAP value in Figure~\ref{fig:GWXGB_features}f illustrates the spatially explicit SHAP value of SVF on the predicted UTCI across different subzones. Red areas with positive SHAP values, mostly concentrated in the city center (P1), northeast residential estates such as Seletar (P2), western industrial areas and ports (P3), and the airport (P4), indicate that areas with large SVF values increase UTCI by a large margin. In contrast, blue areas with negative SHAP value in the Simpang area (N1), the central water catchment and Bukit Timah Hill
(N2), and the Tengah areas (N3) show that the lower SVF tend to reduce UTCI substantially.

\subsection{The nonlinear relationships between urban factors and UTCI}

\begin{figure}[h!]
    \centering    \includegraphics[width=1.0\linewidth]{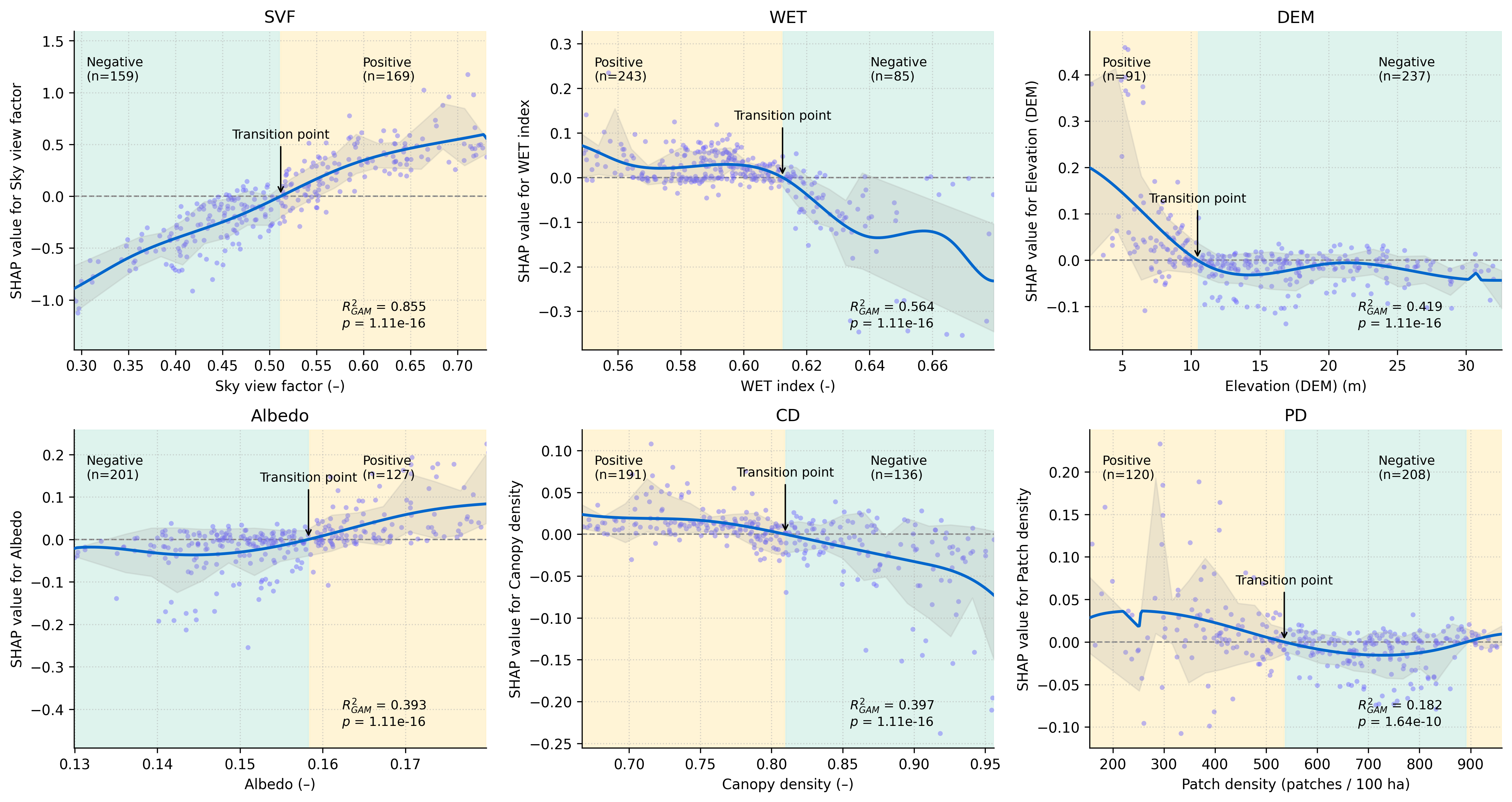}
    \caption{SHAP dependence plots with GAM smoothers showing the non-linear relationships between the six most influential factors and UTCI, including sky view factor (SVF), wetness index (WET), elevation (DEM), albedo, canopy density (CD), and patch density (PD). “Transition point” denotes the first zero-crossing of the smoothed SHAP curve, indicating an approximate transition from negative to positive model-attributed effect (or vice versa).}
    \label{fig:shap_top6}
\end{figure}

SHAP dependence plots coupled with Generalized Additive Model (GAM) smoothers in Figure~\ref{fig:shap_top6} further reveal the nonlinear relationships between the six most influential predictors and UTCI. 
The top-ranked SVF exhibits a strong monotonic increase after a tipping point at approximately 0.51, indicating that enclosed or shaded areas help reduce heat stress by limiting radiative load, whereas beyond this value, greater sky openness amplifies heat stress through enhanced solar exposure. WET exhibits an inverse non-linear effect: UTCI decreases with increasing WET beyond 0.612, reflecting moisture-driven cooling and vegetation influence. DEM also exhibits a cooling association with increasing elevation. The GAM curve begins to decline above approximately 10 m, suggesting that higher terrain may be linked to enhanced airflow, reduced surface heat storage, or other elevation-related microclimatic effects.

\begin{figure}[h!]
    \centering    \includegraphics[width=0.7\linewidth]{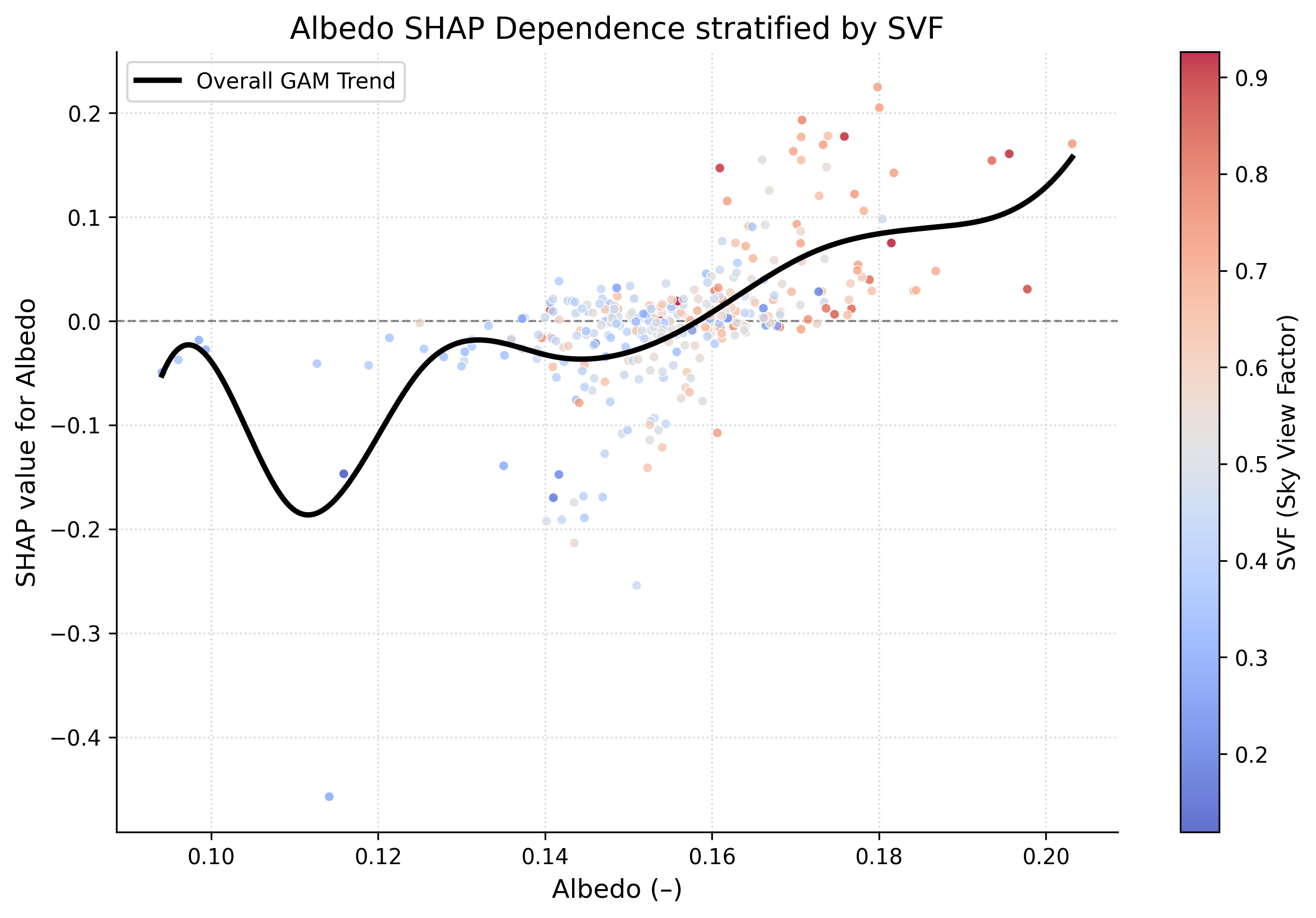}
    \caption{SHAP dependence of Albedo with Sky View Factor (SVF) interaction. Scatter points indicate the SHAP value of Albedo for individual observations, colored by their respective SVF to highlight localized feature interactions. The solid black line represents the overall Generalized Additive Model (GAM) trend, demonstrating a non-linear relationship between albedo and UTCI.}
    \label{fig:shap_albeo}
\end{figure}

Although high albedo could reduce surface and air temperature through increased reflectivity \citep{schneider2023evidence,akbari2012long,prado2005measurement,yi2025sub}, our SHAP–GAM analysis indicates a warming effect with higher albedo values. Stratified SHAP analysis in Figure~\ref{fig:shap_albeo} further indicates that the positive association between albedo and UTCI is concentrated primarily in open/high-SVF environments. Under high-SVF conditions, limited shading increases direct solar radiation receipt, while high-albedo surfaces reflect a greater proportion of incoming shortwave radiation. Although our spatial framework does not directly measure pedestrian-level shortwave flux, these SHAP interaction patterns strongly support the physical hypothesis that high-albedo pavements intensify the combined effects of multiple shortwave radiation reflections, scattering the radiant flux directly onto pedestrians. This hypothesized mechanism aligns with established urban physics, where elevated $T_{mrt}$ subsequently contributes to increased UTCI \citep{schneider2023evidence,schrijvers2016effect,yi2025assessing}. Moreover, previous studies have shown that increased albedo is associated with reduced cloud cover, especially in high urban surface fractions \citep{hamwey2007active,jacobson2012effects}, which is accompanied by increased shortwave radiation.

The SHAP–GAM dependence plot for CD reveals a pronounced non-linear relationship with UTCI. At CD values below the tipping point at approximately $0.81$, SHAP values remain close to zero, partly because much of the shading effect associated with sparse or moderately dense tree cover is already captured by SVF in the model. As a result, CD contributes little additional explanatory power. Beyond the tipping point, however, the dense and more continuous canopy structure leads to substantial reductions in UTCI. The cooling effect strengthens further for CD $>0.85$, reflecting the combined benefits of enhanced shading, reduced $T_{mrt}$, and increased evapotranspirative cooling from cohesive tree canopies.

PD was computed at the landscape level in Fragstats, meaning that all land-cover classes were included when counting patches. Low PD ($<$ 550 patches/100 ha) represents coarse-grained landscapes dominated by large continuous units, which is associated with increased UTCI. As PD increases to moderate levels ($\sim$550–750 patches/100 ha), the landscape becomes more spatially fragmented and finely mixed, resulting in negative SHAP values. However, very high PD ($>$ 900 patches/100 ha) reflects highly fragmented landscapes composed of many small, disconnected patches, which contributes to lower cooling efficiency \citep{qi2025investigating} and a slight increase in UTCI.

Overall, the SHAP–GAM analysis highlights that urban factors exhibit non-linear and threshold-dependent relationships with UTCI. GAM curves combined with SHAP values reveal complex effect directions that cannot be captured by linear models, underscoring the necessity of using interpretable nonlinear modeling to understand and design effective urban-climate mitigation strategies.

\section{Discussion}
\subsection{Discrepancies between LST and UTCI}

The observed discrepancies between LST and UTCI reflect conceptual and physical differences between surface radiative temperatures and human physiological heat stress. Although LST and UTCI exhibit an overall positive but non-linear correlation, they differ systematically in both magnitude and spatial configuration. Compared to UTCI, LST exhibits higher maximum and lower minimum values, indicating a wider thermal range driven by the strong radiative heating and cooling of surface materials and characteristics \citep{small2006comparative}. LST captures the instantaneous radiative temperature of the land surface, which responds sharply to solar heating. In contrast, UTCI integrates air temperature, humidity, wind speed, and radiation to represent pedestrian-level heat stress, which is buffered by atmospheric processes such as convection, radiation exchange, and physiological factors \citep{brode2012deriving,fiala2012utci,jendritzky2012utci,zhan2025satellite}, leading to a more moderated spatial and temporal pattern.

These spatial mismatches occur across land cover types and subzones. In built and impervious areas, UTCI values are systematically lower than LST. This discrepancy is driven by pedestrian-level microclimatic regulation, whereby shading from surrounding buildings substantially reduces shortwave radiation exposure and mean radiant temperature, thereby alleviating human thermal stress \citep{briegel2025satellite}. When combined with the canopy shading and evapotranspirative cooling provided by urban vegetation, these mechanisms effectively mitigate human thermal stress —pedestrian-level benefits that are inherently missed by satellite-derived surface temperature measurements \citep{zhan2025satellite}. Conversely, in open areas with sparse or no canopy cover, such as grassland or barren land, UTCI may remain elevated due to direct solar exposure and limited shading, despite relatively moderate surface temperatures. Water bodies exhibit the lowest LST due to the high specific heat capacity of water, and the water's stable, cooler temperature during hot periods provides a significant local cooling effect on the air directly above it \citep{oke2002boundary}. However, the high humidity and intense solar radiation without shading above the water result in a much higher UTCI value. 

At the subzone level, discrepancies between LST and UTCI are most pronounced in areas characterized by complex 3D urban morphology. These divergences arise from the differential sensitivity of the two metrics to urban morphology and microclimatic regulation. LST can be better explained by built-surface characteristics (NDBI, NDVI, BD), whereas UTCI is more sensitive to morphological and radiative factors (SVF, DEM, and albedo). Reduced SVF, increased building height, and canyon enclosure enhance shading from surrounding buildings, thereby mitigating UTCI even when surface temperatures remain high \citep{briegel2025satellite}. As a result, LST is considerably less sensitive to independent 3D urban morphological features, making it a poor proxy of human-centric heat stress \citep{nazarian2022integrated,tuholske2021global,muse2024daytime,zhan2025satellite}. 

These systematic discrepancies highlight the spatially heterogeneous nature of urban heat exposure. The contrast between these two indices underscores that satellite-derived LST tends to overestimate heat exposure relative to pedestrian-level heat stress in shaded or ventilated environments, particularly within built-up areas \citep{zaerpour2025increasing,zhan2025satellite}. Therefore, we must recognize the potential misleading risks of using LST alone as the primary basis for climate-informed urban planning, particularly in high-density cities. Interventions guided solely by surface temperature may overlook areas of elevated human heat stress and misallocate cooling resources. Integrating human-centric metrics such as UTCI is therefore essential for identifying exposure hotspots and designing effective heat adaptation strategies.

\subsection{Strengths of the explainable GW-XGBoost framework}

The GW-XGBoost framework proves to be a robust tool for this study, successfully balancing high-precision spatial prediction with the ability to quantify complex local interactions. The superior performance of the GW-XGBoost model, as evidenced by the higher $R^2$ (0.905) and lower error metrics (MAE = 0.172~$^\circ$C) compared to global baselines, validates the necessity of accounting for spatial non-stationarity in urban thermal analysis. Unlike traditional global XGBoost algorithms that force a singular, uniform relationship across the entire study area, this geographically weighted framework incorporates spatial heterogeneity through localized weights \citep{fouedjio2024locally,grekousis2025geographical,wang2024geographically,yang2023geographically}. By allowing model parameters to vary across space, GW-XGBoost significantly enhances our understanding of localized variable impacts across highly heterogeneous urban subzones.

To further contextualize the nonlinear relationships identified by the GW-XGBoost and SHAP-GAM framework, we compared our findings with recent studies from Hong Kong, Shenzhen, and Singapore. Overall, our results align with previous evidence that daytime pedestrian heat stress is strongly shaped by radiative exposure, shading, building morphology, and vegetation structure. For example, \citet{jia2025hybrid} found that SVF is strongly associated with outdoor thermal comfort and that open or low-rise urban layouts experience higher heat stress in Hong Kong, supporting our finding that UTCI increases markedly with moderate to high levels of sky exposure. Similarly, evidence from Shenzhen shows that building morphology has competing nonlinear effects: greater built area can increase UTCI through heat storage and reduced ventilation, whereas taller or more enclosed structures can reduce radiant exposure through shading \citep{yang2024optimizing}. This is consistent with our interpretation that vertical morphology affects UTCI mainly through sky exposure and shading geometry. The Singapore-based GPBoost optimization study similarly showed that tree fraction contributes to UTCI reduction, while LAI exhibits a U-shaped nonlinear effect due to the balance among solar interception, canopy heat absorption, and ventilation blockage \citep{chen2026machine}. This finding supports our interpretation that shading and vegetation interact nonlinearly in shaping UTCI, rather than influencing pedestrian-level heat stress independently. A key advancement in our study, however, is the direct comparison between surface-based LST with human-centric UTCI, which demonstrates that LST underrepresents pedestrian-level radiative and shading processes, especially the role of SVF. Overall, these comparisons show that our nonlinear results are consistent with previous UTCI and urban morphology studies, while extending them through a spatially explicit framework that reveals spatially varying driver importance across the city.

\subsection{Implications for climate-informed urban planning}

The discrepancies between LST and UTCI highlight the need to rethink the indicators used in climate-informed urban planning. While LST remains valuable for assessing surface overheating and material performance, it does not reliably represent pedestrian-level heat stress, particularly in dense urban environments where shading, enclosure, and ventilation strongly modulate thermal stress \citep{nazarian2022integrated,tuholske2021global,muse2024daytime,zhan2025satellite}. Therefore, relying on LST alone as the primary basis for planning may misidentify heat-risk hotspots and lead to suboptimal allocation of mitigation measures. For high-density tropical cities such as Singapore, integrating human-centric thermal indicators such as UTCI is essential for identifying where heat is experienced most acutely and for designing more effective adaptation strategies. Rather than advocating for the replacement of LST, our findings support a complementary framework in which LST informs surface-level thermal characteristics, while UTCI captures human-centric heat stress. Together, they provide a more complete representation of urban thermal environments.

Importantly, the spatially explicit SHAP results reveal substantial heterogeneity in the relative importance of heat drivers across subzones, indicating that heat mitigation strategies should be spatially differentiated rather than uniformly applied. For example, in compact residential districts, UTCI is predominantly governed by enclosure-related variables such as SVF, which regulate shortwave exposure, longwave trapping, and wind attenuation. This suggests that heat mitigation in these environments should not simply maximize either openness or greening, but instead balance shading provision with the preservation of local ventilation pathways \citep{li2018mapping,li2024cooling}.

Beyond identifying dominant drivers, the nonlinear response revealed by SHAP-GAM analysis offers subzone-scale, model-informed insights into the intensity of intervention required to achieve meaningful cooling benefits. UTCI in high-density tropical cities is shaped by nonlinear and context-specific processes, rather than by uniformly incremental responses to individual variables. Among these, SVF emerges as the dominant driver of UTCI, with a threshold at approximately 0.51. Below this value, more enclosed or shaded urban configurations help suppress UTCI by limiting radiative exposure \citep{oke2002boundary,nazarian2022integrated}, whereas above this threshold, increasing sky openness is associated with a monotonic rise in pedestrian heat stress \citep{jia2025hybrid}. This finding suggests that open and highly exposed urban spaces should be prioritized for shading-oriented design. Tree canopy density also shows a delayed cooling response: increases below a threshold of approximately 0.81 provide limited benefit, whereas sufficiently dense and continuous canopy yields substantially stronger cooling \citep{li2024cooling}. Consequently, sparse greening fails to provide meaningful thermal relief, underscoring that canopy continuity is as critical as canopy presence for providing additional cooling through combined shading, reduced $T_{mrt}$, and enhanced evapotranspiration \citep{oke2002boundary,shashua2009cooling}. Patch density further demonstrates a nonlinear relationship, indicating that moderate landscape fragmentation (PD of 550–750 patches/100 ha) is more favorable for cooling efficiency, while highly coarse or overly fragmented configurations are associated with higher UTCI, underscoring the importance of spatial configuration alongside land-cover composition \citep{qi2025investigating}. Finally, contrary to conventional expectations, albedo exhibits a net warming effect beyond 0.158 in Singapore’s dense urban context, indicating that reflective materials should be applied cautiously in open pedestrian environments with high SVF, where increased shortwave reflection may elevate radiant heat exposure \citep{schneider2023evidence,schrijvers2016effect,prado2005measurement}. Specifically, the thresholds identified from SHAP-GAM analysis should be interpreted as context-specific, model-based turning points in the fitted nonlinear response curves, rather than fixed planning standards. These values indicate where the marginal effects of urban form variables on LST or UTCI begin to change within the study context. Therefore, they provide exploratory evidence for identifying potentially effective ranges of urban heat mitigation, but should be further validated before being translated into general planning guidelines. Overall, these results suggest that effective heat mitigation in Singapore should be threshold-aware and spatially differentiated: reduce excessive sky exposure in open areas, ensure that greening is sufficiently dense to become effective, avoid both overly coarse and overly fragmented patterns of landscape configuration, and evaluate reflective materials together with pedestrian-level radiation conditions rather than surface temperature alone.

\subsection{Limitations and prospects}

While this study provides insights into the discrepancy between LST and UTCI in Singapore, the generalizability of the findings should be interpreted within specific boundary conditions. First, although the GW-XGBoost framework is generally transferable, the model-derived threshold values and variable importance rankings are likely to be context-specific. The observed patterns are characteristic of tropical, high-density urban environments, where high humidity and intense solar radiation are prevalent under Singapore’s equatorial climate. In such contexts, the decoupling of LST and UTCI is likely a robust phenomenon due to the significant role of $T_{mrt}$ in shaping human thermal sensation. However, in temperate or arid climates, the relationship between LST and UTCI may shift, as surface radiation might dominate the thermal environment differently across seasons. Future research should validate whether the identified driving factors maintain their relative importance in cities with different climatic backgrounds and urban fabrics. Additionally, a limitation is the temporal mismatch between the May 2018–2020 LST composite and the May 2019 UTCI field. This mismatch may affect the absolute magnitude of LST–UTCI differences due to interannual weather variability or short-term surface changes. However, because both datasets represent May daytime conditions and the analysis focuses on relatively stable urban morphology, the main spatial patterns and morphology–heat relationships are expected to be less affected.
Moreover, while this study focused on daytime heat exposure when shortwave radiation and pedestrian vulnerability peak, the absence of nighttime analysis remains a limitation. The physical drivers of the thermal environment shift drastically after sunset. For instance, a low SVF—which provides essential cooling shade during the day—becomes a liability at night by trapping longwave radiation and impeding convective cooling. Evaluating these diurnal contrasts, specifically comparing daytime and nighttime temperatures and their shifting relationships with complex urban morphology \citep{wang2025patterns}, is a key focus of our ongoing research.

Moreover, meteorological data were obtained from NLR with an original spatial resolution of approximately 2 km and subsequently downscaled to 30 m through spatial interpolation to support city-wide analysis. We acknowledge potential biases relative to local meteorological station observations. On one hand, it remains challenging to downscale meteorological variables to a hyperlocal level, particularly given the sparse station network and the lack of sufficient input variables \citep{sun2024deep}. On the other hand, as described in the Methodology, we deliberately refrained from generating interpolated high-resolution surfaces from station data to preserve predictor–response independence. While this decision avoids embedding auxiliary covariate information into the target variable (UTCI), it may also limit the representation of fine-scale microclimatic variability captured by in situ observations. Third, it must be noted that while our UTCI calculation incorporates spatially continuous meteorological data, the native 2-km resolution of the inputs means they function primarily as meteorological forcing fields. These fields (air temperature, relative humidity, and wind speed) provide the background atmospheric state, whereas the $T_{mrt}$ was resolved using fine-scale urban geometry and radiation modeling. We do not claim that these coarse meteorological fields fully capture microclimatic heterogeneity in localized fluid dynamics, such as sensible heat advection or street-canyon wind attenuation. Rather, the modeling framework represents how relatively smooth background meteorological conditions are spatially modulated by local urban form, shading, and radiative exchange, which ultimately generates the fine-scale variation in pedestrian heat stress observed in this study. Future studies could consider integrating multi-source data, including satellite products, reanalysis datasets, dense sensor networks, or mobile measurements, with physics-guided approaches, such as novel Weather Research and Forecasting (WRF) models \citep{gao2025wrfautoml,hong2026wrf}, or uncertainty-constrained data fusion techniques to better capture micro-scale thermal variability while preserving predictor–response independence \citep{wen2025can,kousis2022environmental,liu2017analysis}.

Nevertheless, the high computational cost of SOLWEIG and physically based UTCI simulation models continues to constrain their application at the full city scale, particularly when high spatial resolution, long simulation periods, or multiple scenario analyses are required. These models rely on detailed radiative transfer calculations, complex urban geometry, and fine temporal discretizations, which collectively impose substantial computational and data demands. This limitation has motivated the development of data-driven and hybrid physical-informed neural network (PINN) approaches that aim to bridge the gap between physical realism and computational scalability \citep{shaeri2025multimodal}. By embedding physical constraints such as energy balance relationships and radiative principles within machine-learning architectures, physics-informed neural networks (PINNs) can approximate key microclimatic processes while reducing computational cost \citep{ren2025physics}. Such hybrid frameworks may enable efficient large-scale simulations, uncertainty quantification, and scenario exploration \citep{yi2025planning}, while retaining interpretability and physical consistency. 

\section{Conclusion}
This study provides a morphology-aware comparison between LST and the UTCI for assessing urban thermal conditions in a high-density tropical city. By integrating high-resolution UTCI modeling with explainable global and geographically weighted machine learning approaches, we demonstrate that surface-based temperature metrics and physiologically relevant heat stress indicators convey fundamentally different information about urban thermal environments and their drivers. While LST exhibits strong associations with two-dimensional surface characteristics such as imperviousness and land cover composition, it fails to adequately represent pedestrian-level thermal exposure. In contrast, UTCI is systematically more sensitive to three-dimensional urban morphology. These findings underscore that relying on LST alone may misrepresent heat-risk hotspots and therefore provide an incomplete basis for climate-informed urban planning and heat risk assessment. 

Collectively, this work advances urban climate research by shifting the analytical focus from surface temperatures toward human-centric heat stress metrics. For urban planners and policymakers in Singapore, these findings suggest that effective climate-informed planning in high-density tropical cities requires spatially differentiated interventions that jointly consider shading, canopy continuity, spatial configuration, and radiative conditions at the pedestrian level. In particular, open and highly exposed urban spaces should be prioritized for shading-oriented interventions, greening strategies should consider the dense and continuous canopy rather than sparse vegetation alone, and landscape patterns should avoid both overly coarse and overly fragmented configurations. Reflective materials should also be assessed together with shading design, canyon geometry, and pedestrian-level radiation exposure. Methodologically, this study further demonstrates the value of combining explainable machine learning with geographically weighted modeling to reveal spatially heterogeneous urban climate processes and support more spatially tailored heat mitigation strategies.

\section*{Acknowledgement}

This research is part of  HD4 (Health-Driven Design for Cities) supported by the National Research Foundation Singapore (NRF) under its Campus for Research Excellence and Technological Enterprise (CREATE) programme.

\section*{Declaration of generative AI and AI-assisted technologies in the writing process}

During the preparation of this work, the authors used ChatGPT in order to improve the readability and language of the manuscript. After using this tool, the authors reviewed and edited the content as needed and take full responsibility for the content of the publication.

\appendix
\section{Pearson correlation matrix}
The correlation matrix presents the Pearson correlation matrix among LST, UTCI, and all predictors, which indicates moderate correlations among several predictor groups (e.g., building morphology metrics), but most pairwise correlations remain below $|r|$ = 0.8, suggesting that severe multicollinearity is limited. 

\begin{figure}[h!]
    \centering   \includegraphics[width=0.8\linewidth]{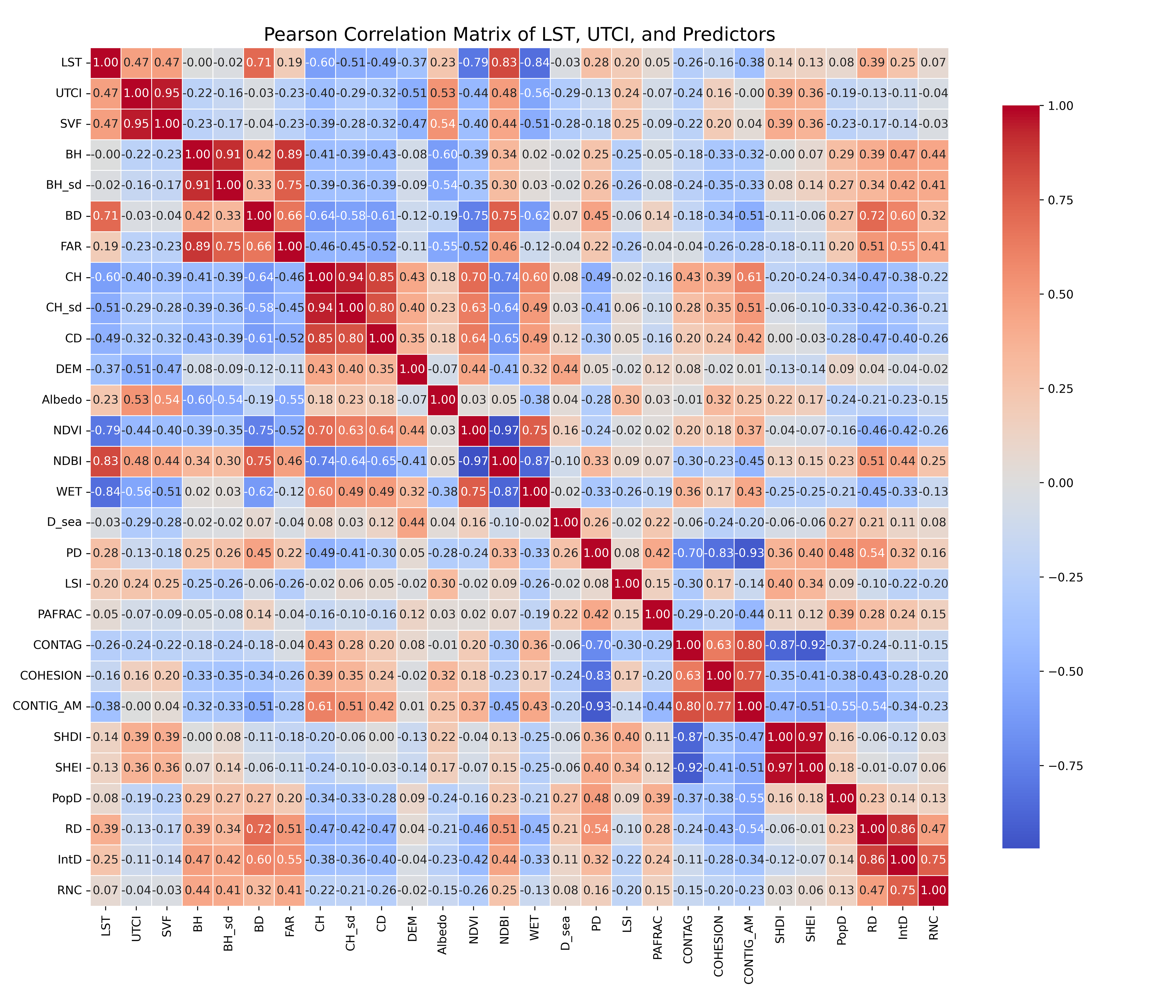}
    \caption{Pearson correlation matrix among LST, UTCI, and the urban factors used in this study. Cell colors indicate the strength and direction of the Pearson correlation coefficients (red = positive correlation; blue = negative correlation.}
    \label{fig:Pearson}
\end{figure}

\section{Real-world comparison of UTCI}

\begin{figure}[h!]
    \centering   \includegraphics[width=\linewidth]{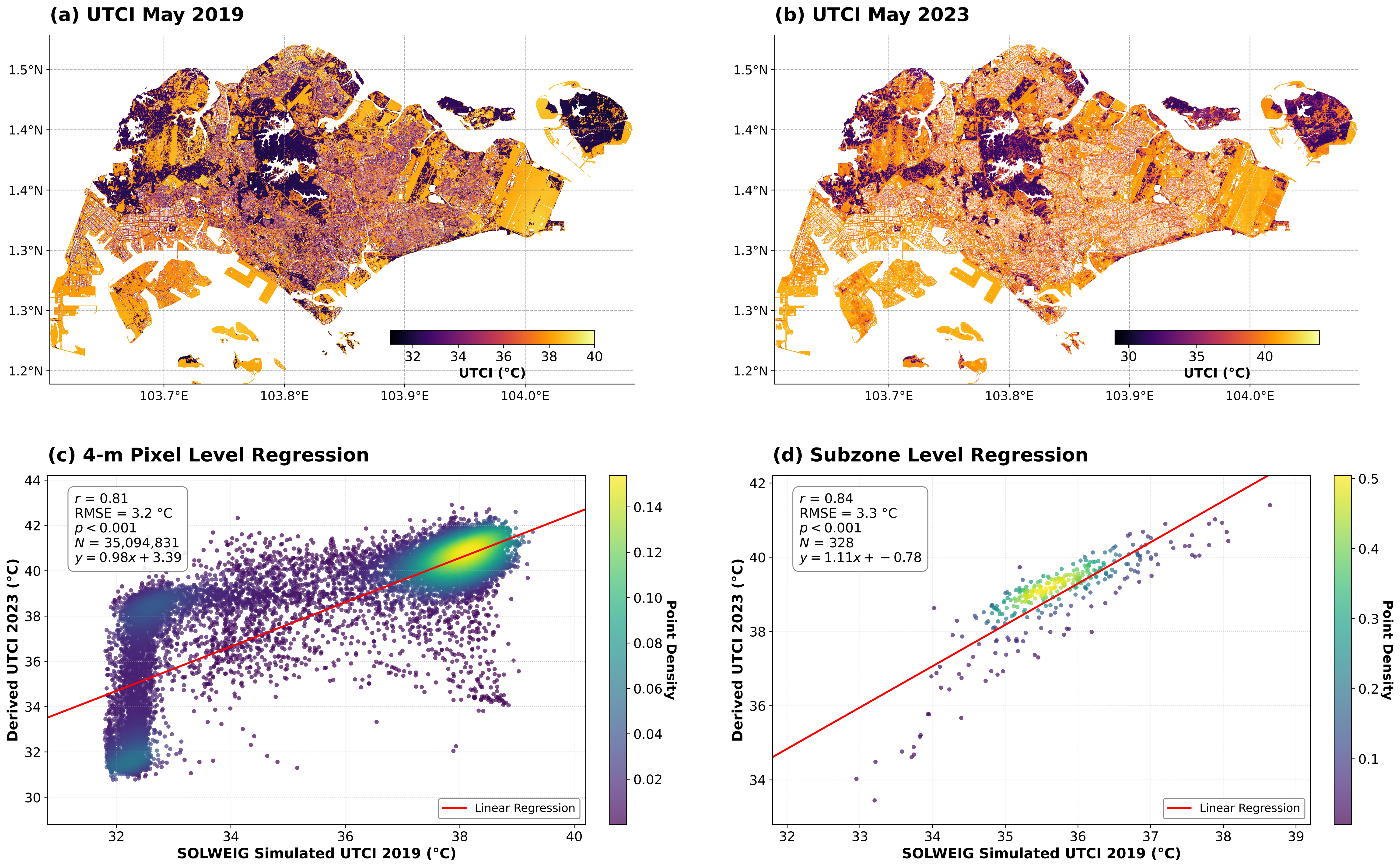}
    \caption{Spatial distribution and statistical comparison of the simulated diurnal UTCI against an independent ENVI-met-based graph neural network microclimate study. Because direct in situ observations were unavailable during the simulation period, the monthly average for May 2019 at 11:00 was selected as the meteorological analogue for the reference day at the exact same time on May 3, 2023. (a) High-resolution (4-m) spatial map of the mean UTCI simulated by the SOLWEIG model for the 2019 analogue, with rooftops and water bodies masked. (b) Corresponding spatial map of the derived UTCI for the 2023 reference, with rooftops and water bodies masked. (c) Pixel-level linear regression and scatter density plot comparing the 2019 SOLWEIG simulations against the 2023 derived estimates ($N = 35,094,831$ pixels). (d) Subzone-level linear regression comparing the aggregated mean UTCI values across 328 urban planning subzones. Both spatial scales demonstrate strong positive correlation ($R = 0.81$ and $R = 0.84$, respectively, $p < 0.001$) with consistent Root Mean Square Errors (RMSE $\approx$ 3.2–3.3 $^\circ$C).}
    \label{fig:UTCI_validation}
\end{figure}

In the absence of direct in-situ UTCI observations for May 2019, we evaluated the plausibility of the simulated UTCI by comparing its magnitude and spatial patterns with an independent Singapore-based study: an ENVI-met-based graph neural network microclimate study that was calibrated and validated against the in-situ sensor network at the National University of Singapore (NUS) on 3 May 2023, one of the extreme heat days in Singapore \citep{xin2026urbangraph,xin2025urbangraph}. Figure~\ref{fig:UTCI_validation} indicates that the SOLWEIG-simulated UTCI for May 2019 successfully reproduces the broad spatial heterogeneity observed in the derived UTCI map for May 2023. Both datasets consistently identify lower heat exposure across vegetated areas, contrasted by elevated UTCI values in dense, open, and coastal-industrial urban zones. Quantitatively, the high-resolution 4-m pixel-level comparison demonstrates a strong positive correlation between the two datasets ($R = 0.81$, RMSE = 3.2 $^\circ$C, $p < 0.001$), based on more than 35 million valid pixels. At the subzone level, this correlation increases slightly ($R = 0.84$, RMSE = 3.3 $^\circ$C), indicating that spatial aggregation reduces local-scale noise while preserving the dominant spatial structure. Although systematic differences remain, particularly the generally higher baseline UTCI values present in the 2023 UTCI dataset, the strong spatial correlations confirm that the 2019 SOLWEIG simulation provides a robust representation of intra-urban UTCI variability, firmly supporting its use for the spatial analysis of heat exposure patterns across Singapore.

\begin{figure}[h!]
    \centering   \includegraphics[width=0.8\linewidth]{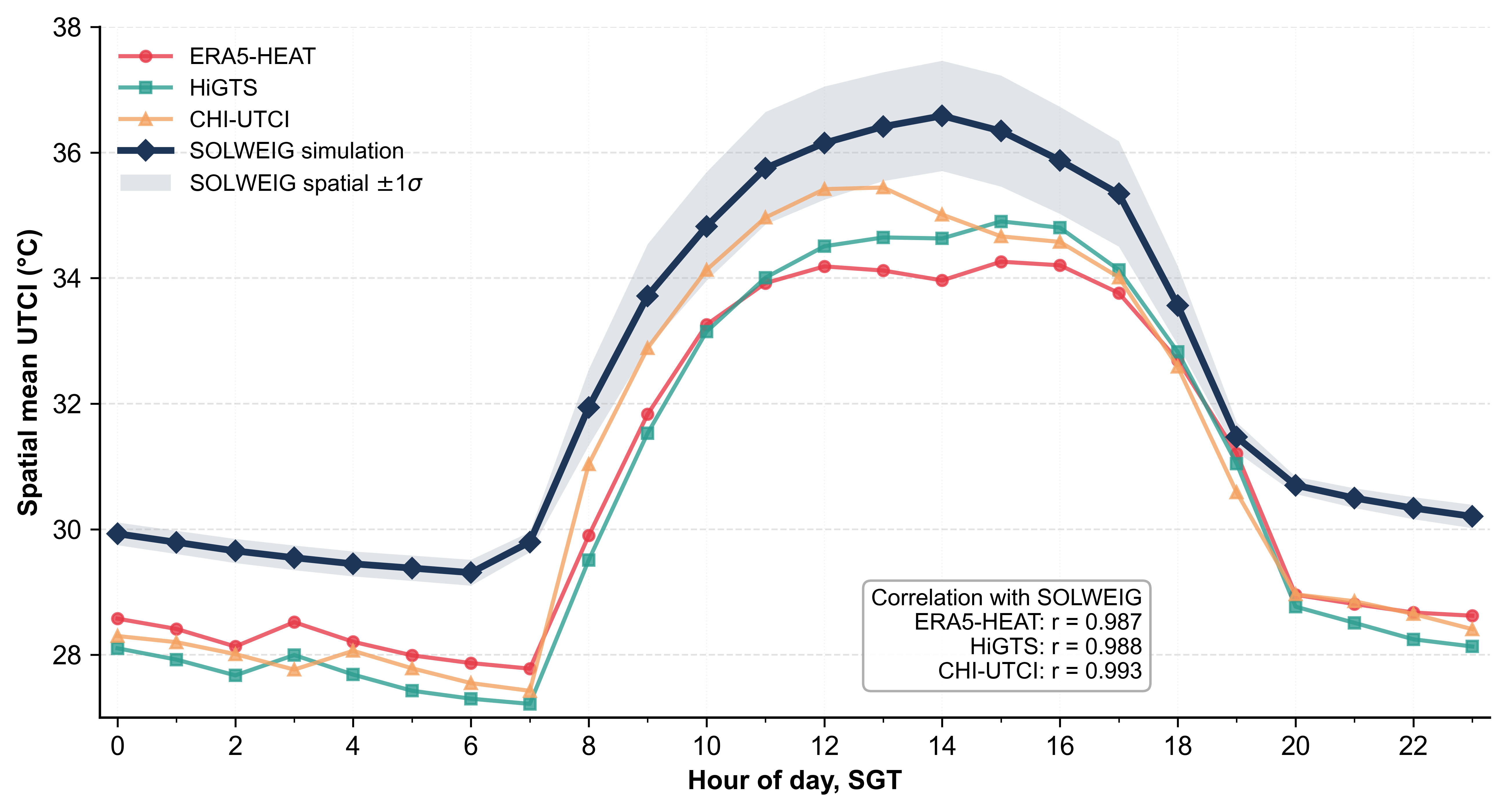}
    \caption{Diurnal comparison between SOLWEIG-simulated spatial mean UTCI and three global UTCI datasets for Singapore in May 2019. The shaded band represents the spatial standard deviation of SOLWEIG-simulated UTCI across subzones. The simulated diurnal profile shows strong agreement with ERA5-HEAT, HiGTS, and CHI-UTCI, with Pearson correlation coefficients of 0.987, 0.988, and 0.993, respectively.}
    \label{fig:UTCI_global}
\end{figure}

To further validate the temporal dynamics of our model, the simulated diurnal UTCI cycle was evaluated against three established global datasets: ERA5-HEAT \citep{di2021era5} ($0.25^\circ \times 0.25^\circ$), HiGTS \citep{jian2024high} ($0.1^\circ \times 0.1^\circ$), and CHI-UTCI \citep{malik2026global} ($0.1^\circ \times 0.1^\circ$). The SOLWEIG-derived 2019 profile (as shown in Figure~\ref{fig:UTCI_global}) accurately captures the diurnal progression of the reference datasets, achieving near-perfect temporal alignment (Pearson's $R = 0.987$, $0.988$, and $0.993$, respectively).

\section{UTCI sensitivity analysis}

A fully systematic perturbation-based assessment of meteorological sensitivity would require a dedicated experimental design involving multiple variables, perturbation ranges, locations, and temporal conditions, and is therefore beyond the main scope of the present study. Furthermore, running the full SOLWEIG spatial model dozens of times can be computationally heavy. Therefore, we added a targeted sensitivity analysis at Changi Airport Meteorological Station. Specifically, a perturbation-based sensitivity analysis was conducted utilizing the meteorological boundary condition at the Changi site ($T_a = 30.0 \text{ }^\circ\text{C}$, $RH = 80.0\%$, $WS = 4.6 \text{ m/s}$, and a SOLWEIG-derived $T_{mrt} = 56.6 \text{ }^\circ\text{C}$) as the baseline. To effectively isolate the sensitivity of the index to ambient thermal and aerodynamic forcing, $T_a$, $T_{mrt}$, $RH$, and $WS$ were independently perturbed within their maximum expected error ranges to quantify their isolated effects on thermal stress. Perturbations were executed across nine evenly spaced intervals: $T_a \in [-2.0, 2.0]\text{ °C}$ at $0.5 \text{ °C}$ steps, $T_{mrt} \in [-6.0, 6.0]\text{ °C}$ at $1.5 \text{ °C}$ steps, $RH \in [-10.0, 10.0]\%$ at $2.5\%$ steps, and $WS \in [-2.0, 2.0]\text{ m/s}$ at $0.5 \text{ m/s}$ steps. For each of the 36 perturbed states (9 steps $\times$ 4 variables), the resulting $UTCI_{perturbed}$ was calculated. The sensitivity of the model was then evaluated by extracting the deviation from the baseline state ($\Delta UTCI = UTCI_{perturbed} - UTCI_{baseline}$). The model was considered robust to an input variable's uncertainty if the resulting $\Delta UTCI$ remained within a threshold of $\pm 1.0 \text{ °C}$, representing a negligible shift in human thermal stress categorization.

\begin{figure}[h!]
    \centering   \includegraphics[width=0.8\linewidth]{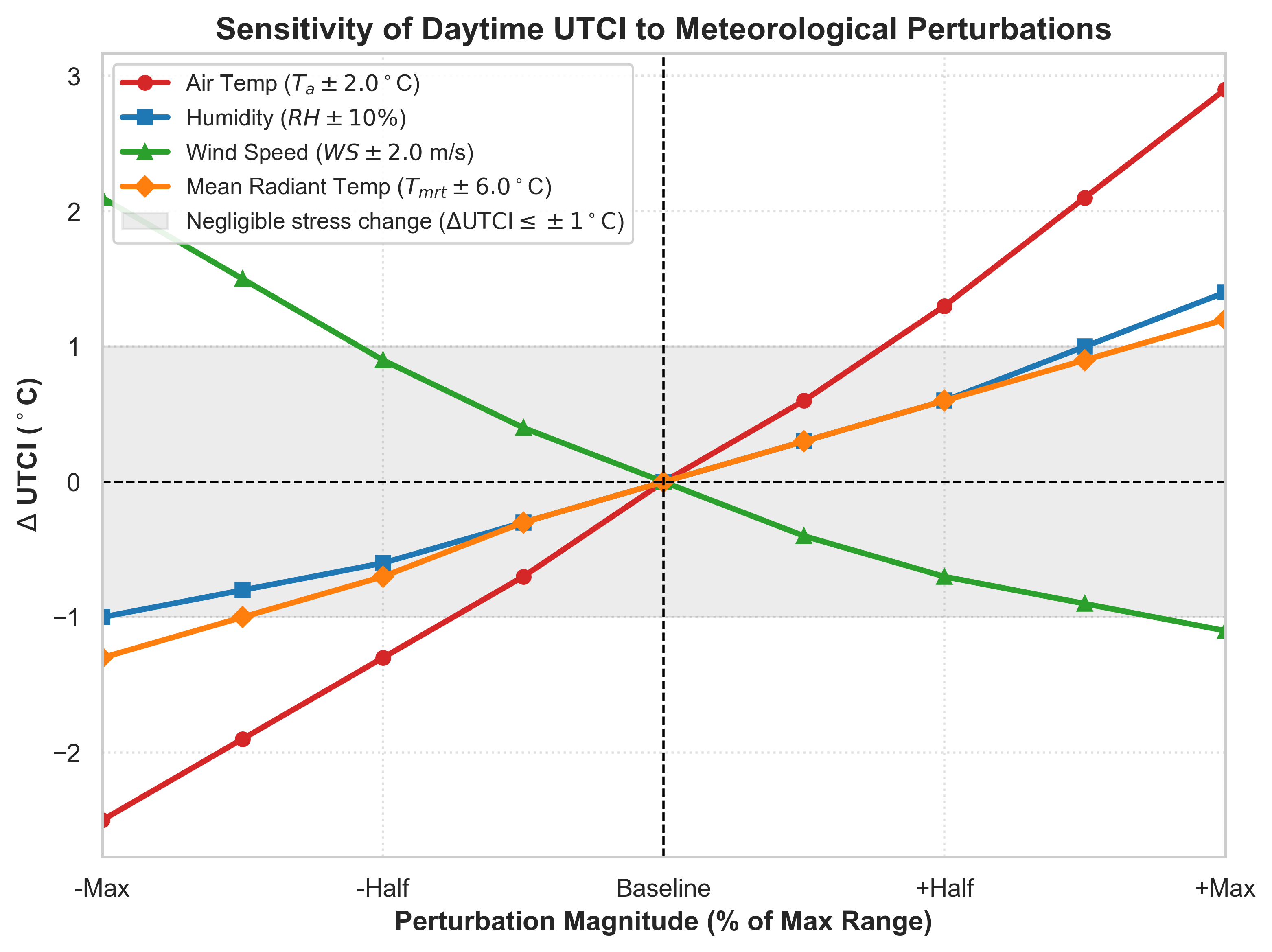}
    \caption{Sensitivity of the simulated daytime UTCI to systematic perturbations in meteorological boundary conditions at Changi Meteorological Station. Input variables include air temperature ($T_a \pm 2.0^\circ$C, red circles), mean radiant temperature ($T_{mrt} \pm 6.0^\circ$C, orange diamonds), relative humidity ($RH \pm 10\%$, blue squares), and wind speed ($WS \pm 2.0$ m/s, green triangles). The shaded grey region denotes the threshold for negligible thermal stress variance ($\Delta\text{UTCI} \le \pm 1^\circ$C).}
    \label{fig:UTCI_sensitivity}
\end{figure}

The results of the sensitivity analysis (Figure~\ref{fig:UTCI_sensitivity}) highlight the UTCI is highly correlated with $T_a$ and moderately sensitive to $T_{mrt}$ and humidity \citep{provenccal2016thermal}. Additionally, $WS$ shows an inverse relationship with UTCI: reduced $WS$ increases UTCI by up to approximately 2.1 $^\circ$C, whereas enhanced $WS$ decreases UTCI by approximately 1.1 $^\circ$C, reflecting the role of ventilation in promoting convective heat loss. Overall, the analysis confirms that uncertainties in $T_a$ can substantially affect modeled UTCI, while most half-range perturbations remain within the negligible-change band of $\Delta\text{UTCI} \le \pm 1^\circ$C). We also acknowledge that this site-level perturbation analysis is not a substitute for a comprehensive city-wide uncertainty assessment. Still, it provides a transparent and reproducible first-order evaluation of meteorological sensitivity.

\section{Spatial autocorrelation of the model residuals}

\begin{figure}[h!]
    \centering   \includegraphics[width=0.8\linewidth]{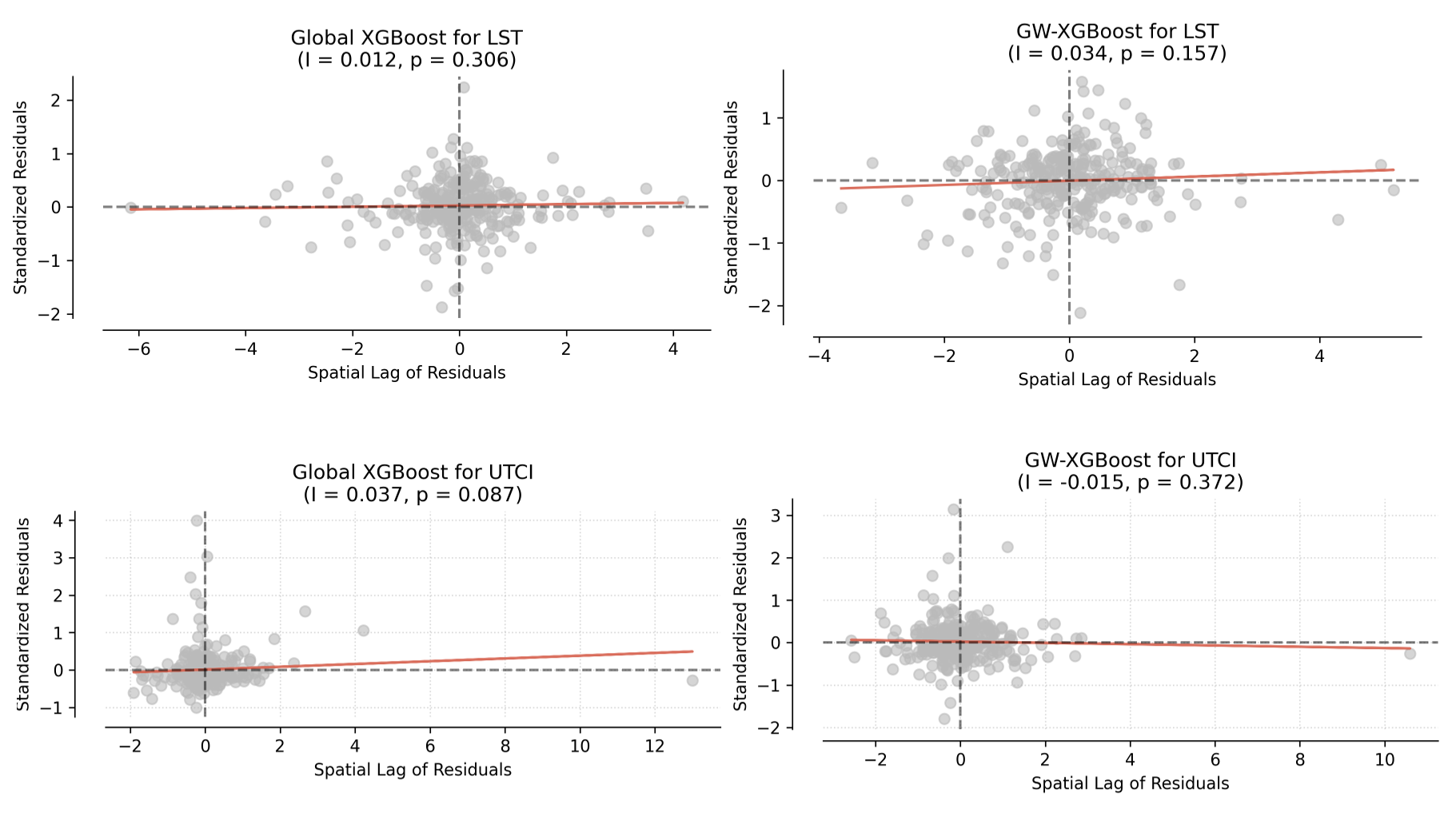}
    \caption{Global Moran's $I$ scatterplots of the residuals for both the Global XGBoost and GW-XGBoost models. A hybrid Queen--nearest-neighbor spatial-weights matrix is used, where isolated subzones are symmetrically connected to their nearest polygon to ensure full spatial connectivity.}
    \label{fig:model_residuals}
\end{figure}

To assess spatial autocorrelation, we calculated the global Moran's $I$ for the residuals of both the global XGBoost and GW-XGBoost models (Figure~\ref{fig:model_residuals}). For LST, the residual autocorrelation was weak and non-significant in both models (Moran's $I = 0.012$, $p = 0.306$ and $I = 0.034$, $p = 0.157$, respectively), indicating that both models already exhibited minimal residual spatial dependence for LST. For UTCI, the global XGBoost residuals showed a marginal tendency toward spatial clustering ($I = 0.037$, $p = 0.087$). However, the implementation of the GW-XGBoost model more effectively accounted for the remaining spatial heterogeneity in UTCI ($I = -0.015$, $p = 0.372$). Overall, although the global models performed well, GW-XGBoost preserved spatial randomness in the residuals while also enabling the identification of highly localized and spatially varying feature relationships.

\section{Meteorological variables and model hyperparameter}

\begin{table}[h!]
\centering
\caption{NSRDB meteorological variables used for UTCI estimation and their temporal handling.}
\label{tab:nsrdb_variables}
\begin{tabular}{p{1.4cm} p{3.2cm} p{1.6cm} p{4.0cm} p{4.0cm}}
\toprule
\textbf{Variable} & \textbf{Description} & \textbf{Unit} & \textbf{Temporal handling} & \textbf{Role in workflow} \\
\midrule
$T_a$ & Air temperature & $^\circ$C & Hourly value at target time & UTCI input \\
RH & Relative humidity & \% & Hourly value at target time & UTCI input \\
WS & 10-m wind speed & m/s & Hourly value at target time & UTCI input \\
GHI & Global horizontal irradiance & W/m$^2$ & Hourly value at target time & Radiation forcing \\
DNI & Direct normal irradiance & W/m$^2$ & Hourly value at target time & Radiation forcing \\
DHI & Diffuse horizontal irradiance & W/m$^2$ & Hourly value at target time & Radiation forcing \\
\bottomrule
\end{tabular}
\end{table}

\begin{table}[htbp]
\centering
\caption{Summary of XGBoost and GW-XGBoost hyperparameter settings, showing the tuned parameters and search ranges used in nested cross-validation, as well as the fixed parameter values applied during model development.}
\label{tab:xgboost_hyperparameters}
\begin{tabular}{llll}
\toprule
\textbf{Parameter} & \textbf{Role} & \textbf{Value / Range} & \textbf{Notes} \\
\midrule
\texttt{n\_estimators}      & tuned & 100, 200, 300, 500 & selected by nested CV \\
\texttt{learning\_rate}     & tuned & 0.1, 0.05, 0.01    & selected by nested CV \\
\texttt{max\_depth}         & tuned & 2, 3, 5, 6          & selected by nested CV \\
\texttt{min\_child\_weight} & fixed & 1                   & preset \\
\texttt{gamma}              & fixed & 0                   & preset \\
\texttt{subsample}          & fixed & 0.8                 & preset \\
\texttt{colsample\_bytree}  & fixed & 0.8                 & preset \\
\texttt{reg\_alpha}         & fixed & 0                   & preset \\
\texttt{reg\_lambda}        & fixed & 1                   & preset \\
\bottomrule
\end{tabular}
\end{table}

\clearpage
\bibliography{bibfile}

\end{document}